\documentclass[final]{Style_File/cvpr}

\usepackage{times}
\usepackage{epsfig}
\usepackage{graphicx}
\usepackage{amsmath}
\usepackage{amssymb}

\usepackage[linesnumbered,ruled]{algorithm2e}
\usepackage{arydshln}
\usepackage{caption}
\usepackage{subcaption}
\usepackage{dblfloatfix}
\usepackage{url}

\usepackage[pagebackref=true,breaklinks=true,letterpaper=true,colorlinks,bookmarks=false]{hyperref}

\def\confYear{CVPR 2021}
\def\cvprPaperID{1346} 

\usepackage{xcolor}

\newcommand\Mark[1]{\textsuperscript#1}

\begin{document}

\title{Content-Aware GAN Compression}

\author{Yuchen Liu\Mark{1},~~~Zhixin Shu\Mark{2},~~~Yijun Li\Mark{2},~~~Zhe Lin\Mark{2},~~~Federico Perazzi\Mark{2},~~~S.Y. Kung\Mark{1}\\
\Mark{1}Princeton University~~~\Mark{2}Adobe Research\\
{\tt\small \Mark{1}\{yl16, kung\}@princeton.edu~~~\Mark{2}\{zshu, yijli, zlin, perazzi\}@adobe.com}
}

\maketitle

\begin{abstract}
Generative adversarial networks (GANs), e.g., StyleGAN2, play a vital role in various image generation and synthesis tasks,
yet their notoriously high computational cost hinders their efficient deployment on edge devices.
Directly applying generic compression approaches yields poor results
on GANs, which motivates a number of recent GAN compression works.
While prior works mainly accelerate conditional GANs, e.g., pix2pix and CycleGAN, compressing state-of-the-art unconditional GANs has rarely been explored and is more challenging.
In this paper, we propose novel approaches for unconditional GAN compression.
We first introduce effective channel pruning and knowledge distillation schemes specialized for unconditional GANs.
We then propose a novel content-aware method to guide the processes of both pruning and distillation.
With content-awareness, we can effectively prune channels that are unimportant to the contents of interest, e.g., human faces, 
and focus our distillation on these regions, which significantly enhances the distillation quality.
On StyleGAN2 and SN-GAN, we achieve a substantial improvement over the state-of-the-art compression method. 
Notably, we reduce the FLOPs of StyleGAN2 by 11$\times$ with visually negligible image quality loss compared to the full-size model.
More interestingly, when applied to various image manipulation tasks, our compressed model forms a smoother and better disentangled latent manifold, making it more effective for image editing.

\end{abstract}
\section{Introduction}
\label{sec:intro}

Generative adversarial networks (GANs)~\cite{goodfellow2014generative} are the leading model for several crucial computer vision tasks like image generation~\cite{brock2018large,karras2020analyzing} and 
image editing~\cite{abdal2019image2stylegan,abdal2020image2stylegan++,harkonen2020ganspace,viazovetskyi2020stylegan2}.
Due to their growing popularity and convincing performance, 
there is an increasing interest in deploying them on edge devices like mobile phones.
However, state-of-the-art GANs often require large storage space, high computational cost, and great memory utility, which disallows them for efficient deployment. For example, StyleGAN2~\cite{karras2020analyzing}
requires 45.1B/74.3B FLOPs to generate a 256px/1024px image, around 150$\times$/250$\times$ more than MobileNet~\cite{sandler2018mobilenetv2}.

A number of network compression techniques have been developed for classification models, including weight quantization~\cite{courbariaux2016binarized,jacob2018quantization},
network pruning~\cite{han2015learning,hu2016network,li2016pruning,zhang2018systematic}, and knowledge distillation~\cite{hinton2015distilling,romero2014fitnets}.
Nonetheless, these methods are not directly applicable for GANs.
For example, although removing channels with low activations~\cite{hu2016network} is effective for classifier compression,  
we find it not better than training a smaller GAN from scratch (Tab.~\ref{tab:metric_effectiveness}).

\begin{figure*}[t]
    \centering \includegraphics[width=0.95\textwidth]{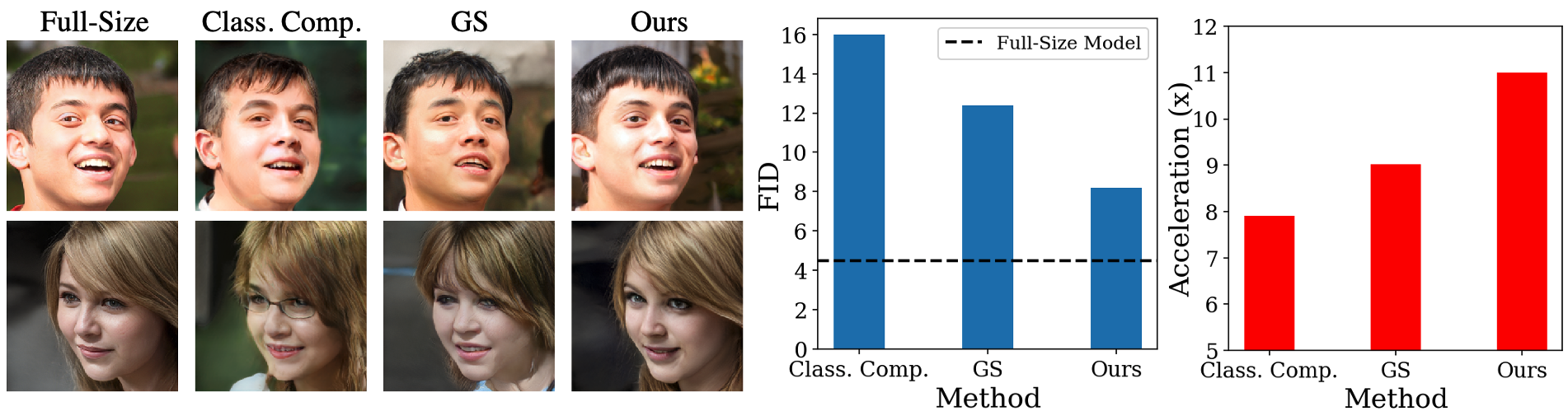}
    \vspace{-0.3cm}
    \caption{
    We demonstrate the advantage of our compression approach on StyleGAN2 over two baseline methods: 
    (1) a conventional classification compression (Class. Comp.) approach with low activation based channel pruning~\cite{hu2016network} and norm-based knowledge distillation~\cite{romero2014fitnets}.
    (2) the state-of-the-art GAN compression method, GAN-Slimming~\cite{wang2020gan} (GS).
    \textbf{Left:} Images generated by full-size model and three compression approaches. Our results show the least artifacts and the best distillation quality.
    \textbf{Right:} Model statistics for three compression schemes. 
    Our model achieves the best FID performance with the highest FLOPs acceleration ratio from the full-size model compared to two baseline methods.
    }
    \label{fig:intro_teaser}
    \vspace{-0.3cm}
\end{figure*}

As such, several specialized GAN compression mechanisms are introduced to learn efficient GAN models 
with the techniques of channel pruning and knowledge distillation~\cite{shu2019co,chen2020distilling,wang2020gan,chang2020tinygan,li2020gan}.
For example, Wang et al.~\cite{wang2020gan} propose GAN-Slimming (GS), which unifies losses of channel pruning and knowledge distillation and achieves the state-of-the-art compression results.
However, these methods
mainly target on conditional GANs (pix2pix~\cite{isola2017image}, CycleGAN~\cite{zhu2017unpaired}, etc.) compression,
and there is little study to compress unconditional GANs (StyleGAN2, etc.).
While conditional GANs normally have paired training data and perform translation from images to images, 
unconditional GANs are trained under completely unpaired setting and have much different source domains (white noises),
which adds extra challenges for the compression.
Therefore, a redesign of channel pruning and knowledge distillation schemes is required for effective unconditional GAN compression.
In addition, these works also miss a significant trait of GANs 
that the output of GANs are images with strong spatial correlation and meaningful semantic contents.
While they prune channels by weights norm or scaling factors and distill images over all spatial locations,
they pay no attention on the generated contents and just treat output images as normal 3D tensors.

To combat these issues, we propose novel approaches to effectively compress unconditional GANs.
We first develop an effective pruning metric to remove redundant channels and explore several distillation losses for unconditional GANs compression.
Different from prior works~\cite{chen2020distilling,chang2020tinygan,li2020gan,wang2020gan}  
where either a norm-based loss or a perceptual loss is used for knowledge distillation, we find that a combination of these losses improves GAN compression results.
With our new pruning and distillation
scheme, we achieve a major improvement in quantitative
measurements over GS.

We then make the first attempt of leveraging the semantic contents in the generated images to guide the GAN compression process of both pruning and distillation.
Specifically, we leverage a content-parsing network to identify contents of interest (COI), a set of spatial locations with salient semantic concepts, within the generated images.
We design a content-aware pruning metric to remove channels that are least sensitive to COI in the generated images.
For knowledge distillation, we focus our distillation region only to COI of the teacher's outputs 
which further enhances target contents' distillation.
The advantage of the content-aware scheme over conventional method is not only demonstrated by a clear improvement in numerical statistics, 
but also visually explained in Fig.~\ref{fig:metric_effectiveness} and~\ref{fig:ca_kd_effectiveness}.
Compared to a classification compression approach and GS,
our compressed model enjoys better generation quality and higher computational acceleration, as shown in Fig.~\ref{fig:intro_teaser}.

Our contributions are four-fold:
(1)~We develop a new framework of channel pruning and knowledge distillation for unconditional GANs compression, 
which achieves a clear improvement over prior methods quantitatively.
(2)~We propose a novel content-aware compression paradigm, 
which leverages generated contents to guide the process of pruning and distillation.
Such a scheme further enhances both visual quality and numerical statistics of the compressed generators.
(3)~Compared to the state-of-the-art GAN compression method, GAN-Slimming~\cite{wang2020gan}, 
our method shows a major advancement in image generation, embedding, and editing on SN-GAN and StyleGAN2. 
(4)~We find that our compressed generators not only have a better resource-performance tradeoff, 
but also own a smoother latent space manifold compared to the uncompressed model, which is beneficial to image editing tasks.


\section{Related Work}

\textbf{Network Compression.} 
To accelerate a classification network, researchers have developed techniques of 
weight quantization~\cite{courbariaux2016binarized,rastegari2016xnor}, 
tensor factorization~\cite{jaderberg2014speeding,lebedev2015speeding}, 
and network pruning~\cite{han2015learning,li2016pruning,hu2016network,liu2017learning,he2018soft,zhang2018systematic,ye2018rethinking,he2019filter}.
Among the network pruning approaches, 
a common method is to remove channels with lower activations~\cite{hu2016network} or smaller incoming weights~\cite{li2016pruning,he2018soft}. 
For instance, 
\cite{hu2016network} removes channels with low activations by averaging their percentage of zeros,
and~\cite{li2016pruning,he2018soft} use the $\ell1$-norm of channels' incoming weights as saliency metric.
However, simply applying~\cite{hu2016network} for GANs pruning would achieve merely the same performance as training from scratch.
Therefore, a more specific approach to identify redundancy in GANs is needed.

\textbf{Knowledge Distillation.}
The idea of knowledge distillation is pioneered by Hinton et al.~\cite{hinton2015distilling} 
to allow a student classifier to mimic the output of its teacher.
Romero et al.~\cite{romero2014fitnets} later propose FitNets which additionally learns from teacher model's intermediate representation.
While norm-based distillation scheme is widely used in distilling recognition models~\cite{yim2017gift,chen2017learning,li2020few,chen2017darkrank}
and has been tried on small GANs~\cite{aguinaldo2019compressing},
applying it on the state-of-the-art unconditional GAN would result in an inferior distillation performance.
Thus, a better design of knowledge distillation is required.

\textbf{Content Awareness.}
In addition to model compression, our work is also related to image saliency/content detection~\cite{zhou2016learning,selvaraju2017grad,hou2017deeply,woo2018cbam,zhao2019pyramid} 
and semantic segmentation~\cite{badrinarayanan2017segnet,yu2018bisenet,fu2019dual}, 
which use deep networks to extract spatial information in images.
Zhou et al.~\cite{zhou2016learning} propose a class activation mapping (CAM) mechanism,
enabling a network to localize class-specific image regions.
This method is generalized in~\cite{selvaraju2017grad} to more network structures and more visual tasks.
Yu et al.~\cite{yu2018bisenet} introduce BiSeNet for image segmentation and is adopted for human face parsing\footnote{https://github.com/zllrunning/face-parsing.PyTorch}.

While the notion of content awareness has been applied for image/video compression~\cite{zund2013content,duan2019content},
it was rarely used under the context of network compression.
Zagoruyko et al.~\cite{zagoruyko2016paying} 
propose an attention transfer scheme to distill the hidden layers of a student classification network.
However, rather than doing a more focused distillation solely on the generated contents as in our proposal, 
they just treat the attention map as an additional feature map and ask the student to match this map in all spatial locations.
While their method is only applied on classifiers' distillation,
our content-aware scheme is designed for more challenging generative models and uses the image contents to guide both processes of pruning and distillation.

\textbf{GAN Compression.}
A number of GAN compression works~\cite{shu2019co,chang2020tinygan,chen2020distilling,li2020gan,wang2020gan,wang2020gan} have been developed to address the issue of efficient GAN deployment, mainly by pruning and distillation.
While~\cite{shu2019co} only uses channel pruning and~\cite{chang2020tinygan,chen2020distilling} use knowledge distillation singly,
\cite{li2020gan,wang2020gan} combine both techniques to enhance compression efficacy.
In particular, Li et al.~\cite{li2020gan} select channels with large incoming weights and distill knowledge with a norm-based loss.
Wang et al.~\cite{wang2020gan} propose a novel GAN-Slimming (GS) approach,
where they impose sparsity constraint on scaling factors for pruning and leverage a style transfer loss~\cite{johnson2016perceptual} for distillation together, with a quantization option.

Although GS achieves state-of-the-art  performance on conditional GAN compression, 
directly applying it to unconditional GANs, such as StyleGAN2, shows sub-optimal results (Tab.~\ref{tab:StyleGAN2_SOTA}).
Hence, 
we propose a different pruning metric as in~\cite{li2020gan,wang2020gan} and a different knowledge distillation scheme, 
which leverages both norm-based loss and perceptual loss for distillation,
rather than using a single loss~\cite{li2020gan,wang2020gan}.
With them, we advance GS on several compression tasks.
We then initiate a novel content-aware compression strategy for both pruning and distillation, 
which further improves model's quantitative measurement.
Not surprisingly, this enhances the visual quality of both generated and edited images, especially for the contents of interest region.
To the best of our knowledge, we are the first one to leverage content awareness in GAN compression.

\section{Methodology}

An unconditional noise-to-image GAN $G$ maps random noises 
from domain $\mathcal{Z}$ to the real world images, $\mathcal{I}$.
We aim to learn a compact and efficient generator $G'$ such that:
(1) their generated images \{$G(z), z\in\mathcal{Z}$\} and \{$G'(z), z\in\mathcal{Z}$\} have similar visual quality;
(2) their embeddings of the real world images \{$Proj(G, I), I\in\mathcal{I}$\}, \{$Proj(G', I), I\in\mathcal{I}$\}  are similar.
Specifically, we leverage the techniques of channel pruning and knowledge distillation, 
where we incorporate content awareness into both processes.

\subsection{Channel Pruning}


\subsubsection{Weight-Based Pruning}

Let $\mathbf{W} \in \mathbb{R}^{n_{in} \times n_{out} \times h \times w}$ denote the convolutional kernel of a layer in $G$ and we aim to quantitate the importance of the $i$th channel, $C_i$. 
Unlike Li et al.~\cite{li2020gan} which uses $\ell1$-norm of $C_i$'s incoming weights, 
we find that $\ell1$-norm of $C_i$'s outgoing weights is a better saliency indicator (shown in Fig.~\ref{fig:metric_effectiveness} and~\ref{fig:metric_FID}), 
and denote the quantity as $\ell1$-out:
\vspace{-0.15cm}
\begin{equation}
\ell1\mathrm{\text{-}out}(C_i) = ||\mathbf{W}_i||_1, \mathbf{W}_i \in  \mathbb{R}^{n_{out} \times h \times w}
\vspace{-0.15cm}
\end{equation}
The channels with a high $\ell1$-out are more informative,
while the ones with lower values are redundant.
 
\begin{figure}[t]
    \centering
    \includegraphics[width=0.48\textwidth]{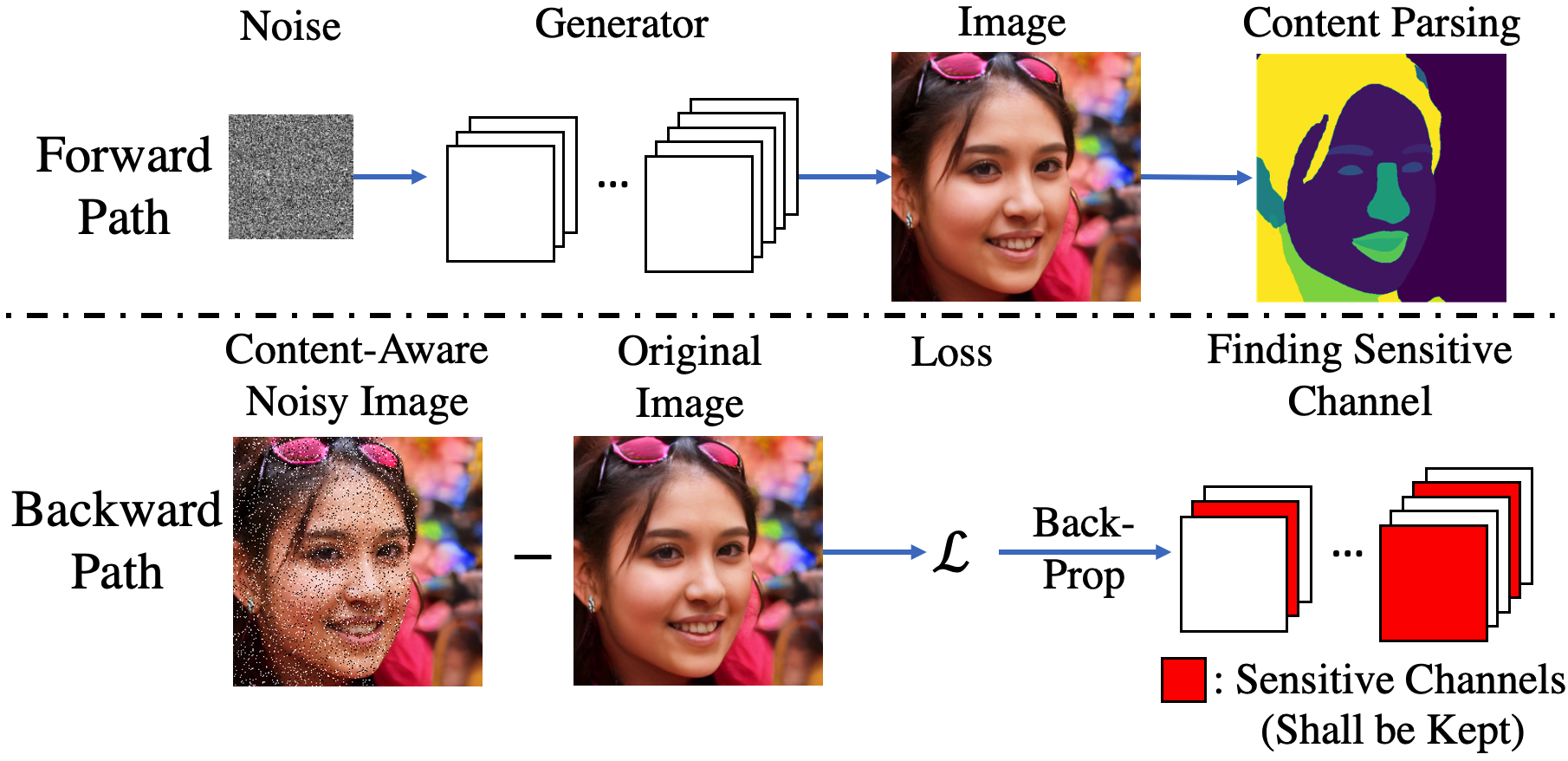}
    \vspace{-0.7cm}
    \caption{
        The content-aware pruning metric with a forward and backward path to identify informative channels.
    }
    \label{fig:Content_Aware_Pruning}
\end{figure}

\vspace{-0.3cm}
\subsubsection{Content-Aware Pruning}

To make the pruned model retain more information on the content of interest, we further develop a content-aware version of $\ell1$-out (named CA-$\ell1$-out),
which is shown in Fig.~\ref{fig:Content_Aware_Pruning} with a forward path and a backward path.

In the forward path, 
we first feedforward a latent variable $z \in \mathcal{Z}$ to $G$ and obtain the generated image $G(z) \in \mathbb{R}^{H \times W \times 3}$.
We then run a content-parsing neural network $Net_p$ on $G(z)$, which returns a content-mask $m \in \mathbb{R}^{H \times W}, m_{h,w} \in \{0, 1\}$, 
where $COI = \{(h,w) | m_{h,w} = 1\}$ denotes the content of interest in the generated image $G(z)$.

For the backward path,
we first add a random image noise $\mathcal{N}$  only on $COI$ of $G(z)$ to obtain a $COI$-noisy images $G_\mathcal{N}(z)$.
A differentiable loss $L_{CA}(G(z), G_\mathcal{N}(z))$ is then constructed 
between the original image $G(z)$  and the $COI$-noisy images $G_\mathcal{N}(z)$.
We then back-propagate $L_{CA}(G(z), G_\mathcal{N}(z))$ to the network's convolution kernel $\mathbf{W}$ and get its gradient $\mathbf{\nabla g} \in \mathbb{R}^{n_{in} \times n_{out} \times h \times w}$.

Such a forward-backward procedure is iterated with multiple samples $z$ to derive the expectation of the content-aware gradient $\mathbb{E}[\mathbf{\nabla g}]$. 
Finally, we measure the $\ell1$-norm of each channel's outgoing filters' gradient as the saliency indicator 
and denote it as CA-$\ell1$-out:
\vspace{-0.1cm}
\begin{equation}
\mathrm{CA\text{-}}\ell1\mathrm{\text{-}out}(C_i) = ||\mathbb{E}[\mathbf{\nabla g}]_i||_1, ~\mathbb{E}[\mathbf{\nabla g}]_i \in  \mathbb{R}^{n_{out} \times h \times w}
\vspace{-0.1cm}
\end{equation}
Intuitively, channels with larger CA-$\ell1$-out are more sensitive to $COI$ of the generated images 
and shall be kept in the pruning process.
Such a content-aware metric selects more informative channels, as shown in Fig.~\ref{fig:metric_effectiveness} and~\ref{fig:metric_FID}.

\subsection{Knowledge Distillation}


\subsubsection{Pixel-Level Distillation}
 
Intuitively, we can impose losses to
reduce the norm-difference between the activations and outputs of $G'$ and $G$.
Based on where the distillation losses are inserted, 
we can categorize the norm-based distillation loss into two types: output only distillation and intermediate distillation. 
For output only distillation, we construct our loss as:
\vspace{-0.2cm}
\begin{equation}~\label{eqn:int_kd}
\mathcal{L}^{norm}_{KD} = \mathbb{E}_{z \in \mathcal{Z}}[||G(z), G'(z)||_1]
\vspace{-0.2cm}
\end{equation}
We can also do intermediate distillation as:
\vspace{-0.2cm}
\begin{equation}
\mathcal{L}^{norm}_{KD} =  \sum^T_{t = 1}\mathbb{E}_{z\in\mathcal{Z}}[||G_{t}(z), f_t(G'_{t}(z))||_1]\text{,}
\vspace{-0.2cm}
\end{equation}
where $G_{t}(z)$ /$G'_{t}(z)$ are the intermediate activations of layer $t$ and $G_T(z)$/$G'_T(z)$ are the output images. 
$f_t$ is a linear transform to match the depth dimension of the activations.

\subsubsection{Image-Level Distillation}

Apart from learning low-level details from the teacher, 
we also want the student to generate perceptually similar outputs.
To achieve this, we adopt neural network based perceptual metrics.
Unlike GAN-Slimming~\cite{wang2020gan}, which uses a style transfer loss~\cite{johnson2016perceptual} for distillation,
we propose to use LPIPS~\cite{zhang2018unreasonable} as our perceptual distillation loss,
which measures the perceptual distance between output images from two generators:
\vspace{-0.15cm}
\begin{equation}
\mathcal{L}^{per}_{KD} =  \mathbb{E}_{z\in\mathcal{Z}}[LPIPS(G(z), G'(z))]
\vspace{-0.15cm}
\end{equation}
In our experiments, we find that LPIPS is a better than~\cite{johnson2016perceptual} for unconditional GAN compression.

\subsubsection{Content-Aware Distillation}

\begin{figure}[t]
    \centering
    \includegraphics[width=0.47\textwidth]{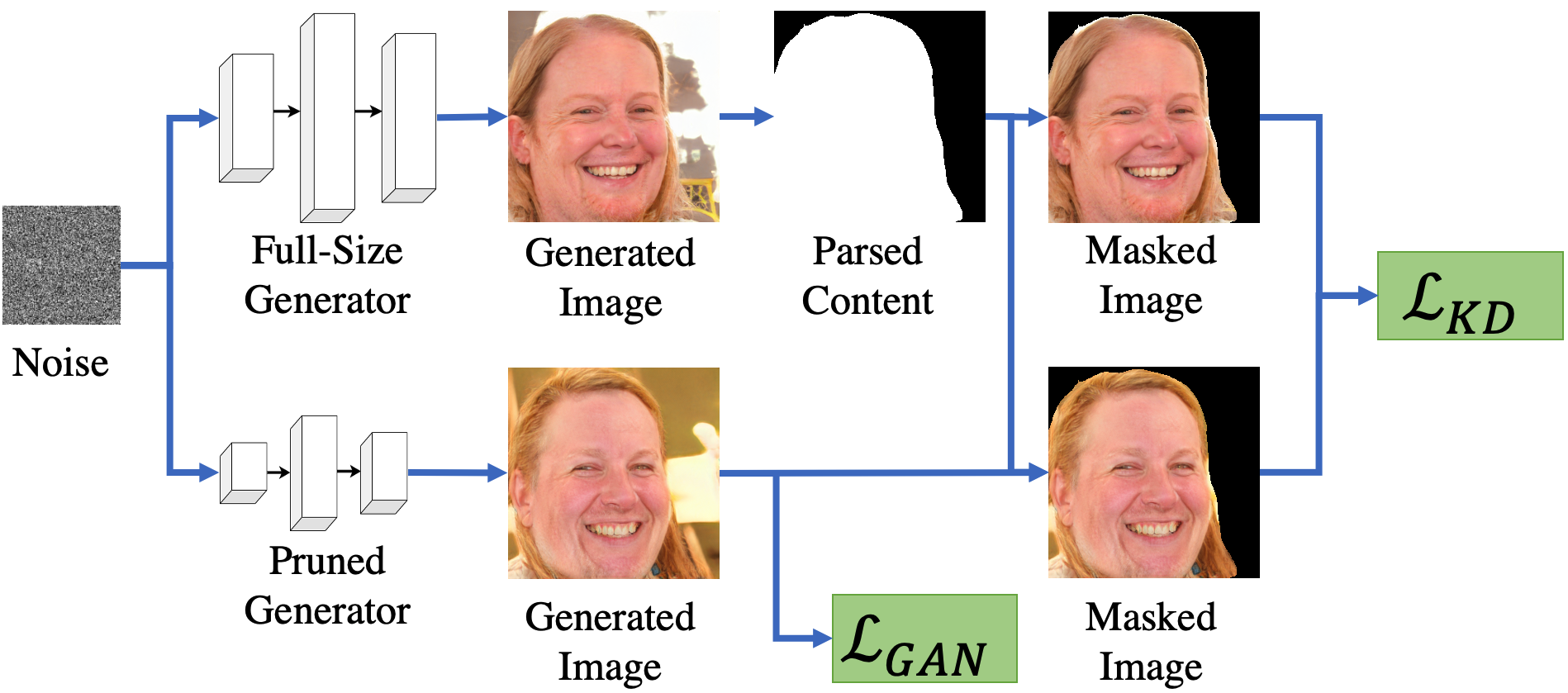}
    \vspace{-0.2cm}
    \caption{
        Our content-aware knowledge distillation (CA-KD) scheme, where the minmax GAN loss is applied on the pruned model's generated images 
        while the knowledge distillation loss is imposed on the content-masked images from the full-size teacher and pruned student. 
    }
    
    \label{fig:Content_Aware_KD}
\end{figure}
Similar to pruning, we introduce content-awareness to knowledge distillation as shown in Fig.~\ref{fig:Content_Aware_KD}, where we focus our distillation on specific contents.
We first feedforward a latent variable $z$ to obtain both networks' generated images, $G(z)$ and $G'(z)$.
Then, we run the content-parsing network $Net_p$ on $G(z)$ and get its content mask $m$.
Based on $COI$ of $m$, we compute two masked images, $G(z)_m = G(z)\odot m$ and 
$G'(z)_m = G'(z)\odot m$, 
where $\odot$ denotes the element-wise multiplication in images' spatial domain.
Under this scheme, 
$\mathcal{L}_{GAN}$ will be imposed on the unmasked generated image of the pruned network, $G'(z)$, 
while the distillation loss for the output images are measured between the masked images $G(z)_m$ and $G'(z)_m$.
Such a content-aware scheme allows a more attentive distillation for the generated contents,
as evidenced in Fig.~\ref{fig:ca_kd_effectiveness}.

\subsubsection{Training Objectives}
In summary, our training loss for the pruned generator $G'$ can finally be formulated as:
\vspace{-0.2cm}
\begin{equation}
\mathcal{L} = \mathcal{L}_{GAN} + \lambda\mathcal{L}^{norm}_{KD} + \gamma\mathcal{L}^{per}_{KD}
\vspace{-0.2cm}
\end{equation}
where $\mathcal{L}_{GAN}$  is the minmax objective for GAN training, 
and $\lambda$ and $\gamma$ are the weights for the knowledge distillation losses.
Unlike prior works~\cite{li2020gan,wang2020gan} singly uses either pixel-level distillation loss $\mathcal{L}^{norm}_{KD}$
or image-level distillation loss $\mathcal{L}^{per}_{KD}$,
we find that it is necessary to combine both of them for an enhanced distillation performance.
Moreover, applying $\mathcal{L}^{norm}_{KD}$ and $\mathcal{L}^{per}_{KD}$ under the content-aware scheme further improves distillation quality.

For $\mathcal{L}_{GAN}$, we derive the student generator $G'$ by pruning $G$, 
while initializing the student discriminator $D'$ with the same architecture and pre-trained weights as teacher discriminator $D$. 
We fine-tune both $G'$ and $D'$ by a standard minmax optimization scheme~\cite{goodfellow2014generative}.

\section{Experimental Results}\label{experiment}

We carry out compression experiments on  models with different computation budgets and datasets with diverse image resolutions
to show the general effectiveness of our approach.
Specifically, we investigate into the following tasks:
SN-GAN~\cite{miyato2018spectral} on CIFAR-10~\cite{krizhevsky2009learning} at 32px,
StyleGAN2~\cite{karras2020analyzing} 
on FFHQ dataset~\cite{karras2019style} at 256px and at 1024px.

\subsection{Evaluation Metrics\label{sec:image_projection}}

We use the following five quantitative metrics to evaluate the image generation and image projection performance of a GAN:
Inception Score (IS)~\cite{salimans2016improved},
Fréchet Inception Distance (FID)~\cite{heusel2017gans}, 
Perceptual Path Length (PPL)~\cite{karras2019style},
and PSNR/LPIPS between real and projected images.

\textbf{Inception Score.}\footnote{We use https://github.com/tsc2017/Inception-Score}
IS is proposed to measure the classification quality of the generated images. 
Specifically, it awards high scores to a generator whose generated images 
could be classified by an inception classifier~\cite{szegedy2016rethinking} with high confidence 
while having a diverse label distribution.  

\textbf{Fréchet Inception Distance.}\footnote{We use https://github.com/mseitzer/pytorch-fid}
FID quantitates the similarity between the synthetic images from a generator and the real-world images.
It is computed by feed-forwarding two sets of images to an inception network
followed by a Fréchet Distance~\cite{frechet1957distance} measurement between their corresponding activation features.

\textbf{Perceptual Path Length.}
PPL is proposed to measure the smoothness of a StyleGAN's latent space~\cite{karras2019style}.
It is derived by calculating the LPIPS distance between two images generated by a pair of little-perturbed latent codes.
Our PPL implementation has two slight modifications from~\cite{karras2019style}:
(1) instead of using a reimplemented LPIPS in TensorFlow~\cite{abadi2016tensorflow}, 
we use the original LPIPS implementation\footnote{https://github.com/richzhang/PerceptualSimilarity};
(2) we fix the perturbation factor $\epsilon=10^{-4}$, and drop the scaling term $\frac{1}{\epsilon^2}$ to make the score more intuitive.

\begin{figure}[t]
    \centering
    \includegraphics[width=0.48\textwidth]{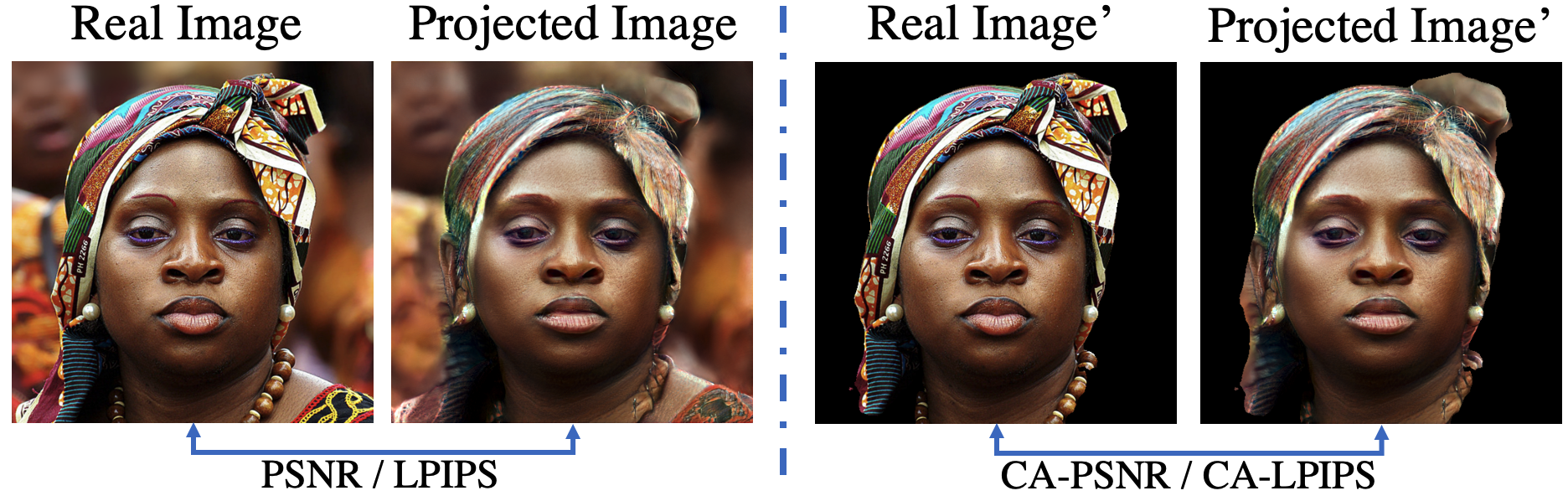}
    \vspace{-0.7cm}
    \caption{Image projection evaluation. We measure the PSNR/LPIPS for the pair of real and projected images, 
    as well as their content-aware masked images.}
    \label{fig:ca_projection}
\end{figure}

\textbf{Image Projection.}
Our image projection evaluation method is shown in Fig.~\ref{fig:ca_projection}.
We sample 55 real human face images from Helen~\cite{le2012interactive} (not appeared in any training dataset) with various lighting conditions, genders, ethnicities, and ages, 
and name the dataset as Helen-Set55, shown in the Supplementary. 
We run an L-BFGS optimizer~\cite{liu1989limited} with 200 iterations to find a real image's latent code in a StyleGAN2.
We feed-forward this latent code and obtain a projected image.
We average LPIPS and PSNR for all pairs of Helen-Set55 images and projected images.

Moreover, in many image editing applications~\cite{adobe_photoshop}, 
we mostly care about the projection quality of the content of interest region. 
Thus, we further propose a content-aware projection evaluation scheme,
where we measure the PSNR and LPIPS between the content-aware masked real and projected images, denoted as CA-PSNR and CA-LPIPS.

\subsection{Pruning Effectiveness}

We first examine the effectiveness of our channel pruning metric on a 256px StyleGAN2.
Given a pretrained generator, we uniformly remove 30\% of channels from each layer by $\ell1$-out or other metrics.
Three baselines are included to show our effectiveness: 
(1) training from scratch by keeping the pruned network structure while re-initializing the weights;
(2) a conventional classification pruning method, 
to remove channels with low activations\footnote{Rather than counting average percentage of zeros,
we choose to measure activations' $\ell1$-norm to remove low activation maps, 
as the activation function in StyleGAN2 is leaky ReLU, not ReLU.}~\cite{hu2016network};
(3) random pruning.
These pruned networks are then fine-tuned by the vanilla training loss, $\mathcal{L}_{GAN}$, where the discriminators see 2.9M real examples in total.

\begin{table}[t]
\fontsize{8.5}{10}\selectfont
\centering

\begin{tabular}{|c|c|c|c|}
\hline
Model & Image Size & FLOPs & FID ($\downarrow$) \\
\hline\hline
 
Original Full-Size & 256 & 45.1B & 4.5 \\
\hline
\multicolumn{4}{|c|}{Compressed Models} \\
\hline

Low-Act Pruning~\cite{hu2016network} & 256 & 22.3B & 7.9\\

Training from Scratch & 256 & 22.3B & 8.1 \\

Random Pruning & 256 & 22.3B & 6.2 \\

$\ell1$-out Pruning & 256 & 22.3B & \textbf{5.4} \\

\hline

\end{tabular}

\vspace{-0.2cm}
\caption{Pruning metric effectiveness investigation. We use FID (the lower the better) to compare $\ell1$-out pruning with three other baselines.}
\label{tab:metric_effectiveness}
\end{table}
\begin{table*}[t]

\fontsize{8.5}{10}\selectfont

\centering

\begin{tabular}{|c|c||c|c||c|c||c|c|c|c|}
\hline
Model & Image Size & $\mathcal{L}^{norm}_{KD}$ Type& $\lambda$ & $\mathcal{L}^{per}_{KD}$ Type & $\gamma$  & FLOPs & FID ($\downarrow$) & PSNR ($\uparrow$) & LPIPS ($\downarrow$) \\
\hline\hline
 
 Original Full-Size & 256 & - & - & - & - & 45.1B & 4.5  & 32.02 & 0.113 \\
\hline
\multicolumn{10}{|c|}{Compressed Models} \\
\hline

 No KD & 256 & - & - & - & - & 1.9B & 15.1 & 30.88 & 0.182 \\

$\mathcal{L}^{norm}_{KD}$ Only & 256 & Output Only & 3 & - & - & 1.9B & 14.2  & 30.79 & 0.186 \\

 $\mathcal{L}^{norm}_{KD}$ Only & 256 & Intermediate & 3 & - & - & 1.9B & 14.4 & 30.78 & 0.187 \\

 $\mathcal{L}^{per}_{KD}$ Only & 256 & - & - & VGG & 3 & 1.9B & 14.6  & 30.81 & 0.186 \\

 $\mathcal{L}^{per}_{KD}$ Only & 256 & - & - & LPIPS & 3 & 1.9B & 13.6  & 30.91 & 0.177 \\
 
$\mathcal{L}^{norm}_{KD}$ + $\mathcal{L}^{per}_{KD}$ & 256 & Intermediate & 3 & LPIPS & 3 & 1.9B &  13.6 & 30.76 & 0.180 \\

$\mathcal{L}^{norm}_{KD}$ + $\mathcal{L}^{per}_{KD}$ & 256 & Output Only & 3 & LPIPS & 3 & 1.9B & \textbf{12.5}  & \textbf{31.03} & \textbf{0.170} \\

\hline
\end{tabular}

\vspace{-0.3cm}
\caption{Results of different knowledge distillation schemes.}
\label{tab:kd_results_96}
\vspace{-0.4cm}
\end{table*}

As shown in Tab.~\ref{tab:metric_effectiveness},  the low-act pruned model
has merely the same FID as training from scratch, even worse than random pruning.
This indicates that directly applying classification pruning metric can fail on GAN compression. 
Moreover, we find that the $\ell1$-out pruned generator
has an only 0.9 FID loss from the full-size model with 50\% less FLOPs,
and it achieves the best FID among compared methods.

\subsection{Knowledge Distillation Schemes}\label{sec:kd_schemes}

We then analyze the effectiveness of different knowledge distillation losses under a high acceleration ratio.
We use $\ell1$-out to uniformly remove 80\% of channels from each layer of a full-size StyleGAN2, 
resulting in an 25$\times$ FLOPs-accelerated generator.
We then retrain the pruned model with 4.3M real examples by six combinations of distillation losses,
which are specified by $\lambda, \gamma$, and the type of norm-based distillation.
Moreover, we include the VGG style transfer loss~\cite{johnson2016perceptual} used in GAN-Slimming distillation~\cite{wang2020gan} for our comparison.
For intermediate distillation, 
we choose the outputs of the $to\_rgb$ modules to construct the loss in Eqn.~\ref{eqn:int_kd}.
As the depth dimension of $to\_rgb$ outputs are always 3, 
we can fix $f_t(x) = x$.

As shown in Tab.~\ref{tab:kd_results_96}, 
the VGG style transfer loss yields inferior results to LPIPS which suggests the need for redesign distillation loss for unconditional GANs compression.
Moreover, the output only $\mathcal{L}^{norm}_{KD}$ + LPIPS $\mathcal{L}^{per}_{KD}$  distillation scheme achieves the best quantitative results for both image generation projection:
(1) it achieves the best FID score of 12.5 which has an FID improvement of 2.6 over the no KD scheme.
(2) it achieves the best image projection results with a PSNR of 31.03 and LPIPS of 0.170.
Such results are interesting that:
(1) rather than prior work which only adopts a single loss distillation scheme~\cite{li2020gan,wang2020gan},
we find that it is necessary to distill knowledge both at the pixel-level and image-level for enhanced results;
(2) while distilling the knowledge in the intermediate features improves conditional GAN's performance~\cite{li2020gan},
it does not help for the case of unconditional GAN like StyleGAN2.

We regard the output only $\mathcal{L}^{norm}_{KD}$ + LPIPS $\mathcal{L}^{per}_{KD}$ as our best KD loss which are used in the following experiments.

\subsection{Comparison to the State of the Art}\label{sec:sota_comparison}

\begin{table}
\fontsize{8.5}{10}\selectfont

\centering

\begin{tabular}{|c|c||c|c||c|}
\hline
Model & Img Size & FLOPs & Param. & IS ($\uparrow$)\\
\hline\hline
 
 Ori. Full-Size & 32 & 1.60B & 4.27M & 8.37$\pm$0.11  \\
\hline
\multicolumn{5}{|c|}{Compressed Models} \\
\hline
GS~\cite{wang2020gan} & 32  & 1.11B & 3.38M & 8.01 \\

Ours & 32  & 0.83B & 2.23M & 8.36$\pm$0.12 \\

Ours-CA  & 32  & 0.83B & 2.23M & \textbf{8.36$\pm$0.08}
\\

\hline

GS~\cite{wang2020gan} & 32  & 0.51B & 2.19M & 7.65 \\

Ours & 32  & 0.42B & 1.20M & 8.21$\pm$0.11 \\

Ours-CA  & 32  & 0.42B & 1.20M & \textbf{8.31$\pm$0.08} \\

\hline
\end{tabular}

\vspace{-0.3cm}
\caption{Comparison to the state-of-the-art method, GAN-Slimming, with SN-GAN on CIFAR-10.}
\label{tab:SN_GAN_SOTA}

\end{table}

\begin{table*}[t]

\fontsize{8.5}{10}\selectfont

\centering

\begin{tabular}{|c|c||c|c|c|c|c||c|c|}
\hline
Model & Image Size & FLOPs & FID($\downarrow$) & PPL($\downarrow$) & PSNR($\uparrow$) & LPIPS($\downarrow$) & CA-PSNR($\uparrow$) & CA-LPIPS($\downarrow$) \\
\hline\hline
 
 Original Full-Size & 256 & 45.1B & 4.5 & 0.162 & 32.02 & 0.113 & 33.03 & 0.076 \\
\hline
\multicolumn{9}{|c|}{Compressed Models} \\
\hline

GS~\cite{wang2020gan} (Our Impl.) & 256  & 5.0B & 12.4 & 0.313 & 31.02 & 0.177 & 32.39 & 0.117 \\

Ours & 256  & 4.1B & 8.9 & 0.145 & 31.37 & 0.149 & 32.67 & 0.099 \\

Ours-CA  & 256 & 4.1B & \textbf{7.9} & \textbf{0.143} & \textbf{31.41} & \textbf{0.144} & \textbf{32.75} & \textbf{0.096} \\

\hline
\multicolumn{9}{c}{}\\
\hline

 Original Full-Size & 1024 & 74.3B & 2.7 & 0.162 & 31.38  & 0.149 & 32.67 & 0.096\\

\hline
\multicolumn{9}{|c|}{Compressed Models} \\
\hline

GS~\cite{wang2020gan} (Our Impl.) & 1024  & 23.9B & 10.1 & 0.211 & 30.74 & 0.189 & 32.17 & 0.121 \\

Ours & 1024  & 7.0B & 8.1 & 0.157 & 30.94 & 0.174 & 32.31 & 0.113 \\

Ours-CA  & 1024 & 7.0B & \textbf{7.6} & \textbf{0.157} & \textbf{30.96} & \textbf{0.170} & \textbf{32.33} & \textbf{0.111} \\

\hline
\end{tabular}

\vspace{-0.3cm}
\caption{Comparison to the state-of-the-art method, GAN-Slimming, with 256px/1024px StyleGAN2 on FFHQ.}
\label{tab:StyleGAN2_SOTA}
\vspace{-0.2cm}

\end{table*}

To further demonstrate the effectiveness of our approach, 
we compare our scheme with the state-of-the-art GAN compression method, 
GAN Slimming (GS)~\cite{wang2020gan}. 
The results are summarized in Tab.~\ref{tab:SN_GAN_SOTA} and~\ref{tab:StyleGAN2_SOTA}, where we consistently outperform GS in all measurements.

\vspace{-0.15cm}
\subsubsection{Experimental Settings}
We compare compressed models by three different approaches: \textbf{Ours}, \textbf{Ours-CA} and \textbf{GS}.

\textbf{Ours.}
We use $\ell1$-out for channel pruning with the best KD loss in Sec.~\ref{sec:kd_schemes} for fine-tuning. 
We set $\lambda = 3$ and $\gamma = 3$.

\textbf{Ours-CA.}
We use CA-$\ell1$-out as the pruning metric
and apply the best KD loss with content-aware distillation (Fig.~\ref{fig:Content_Aware_KD}).
On CIFAR-10, we use the class activation mapping (CAM)~\cite{zhou2016learning} to 
detect generated images' $COI$.
For FFHQ, 
we use a BiSetNet~\cite{yu2018bisenet} to parse human faces.
The $COI$ is the entire human face, i.e., excluding the clothes and the image background.
For CA-$\ell1$-out, 
we use the salt-and-pepper noise as $\mathcal{N}$, and use an $\ell1$-loss for $L_{CA}$.
For content-aware distillation, we use the same value for $\lambda$ and $\gamma$.

\textbf{GS.} 
We use the numbers reported in the original paper for the SN-GAN comparison.
As GS has not been applied to StyleGAN2 in the previous work, 
we make our own implementation following their three-step method:
(1) fine-tune the full-size network with a combination of a minmax GAN loss, an $\ell$1 sparsity loss on scaling factors, and a VGG distillation loss~\cite{johnson2016perceptual};
(2) remove channels with zero scaling factors in the tuned network;
(3) fine-tune the pruned network with the GAN loss and the VGG style transfer loss.

On SN-GAN, our discriminator sees 1.6M images in the fine-tuning process.
On StyleGAN2, the discriminators in both our scheme and \textbf{GS} see the same number of images, 7.5M, 
ensuring the fairness for the comparison. 
Extra experiment details are included in the Supplementary.

\vspace{-0.15cm}
\subsubsection{Results}
\begin{figure*}[t]
    \centering \includegraphics[width=0.85\textwidth]{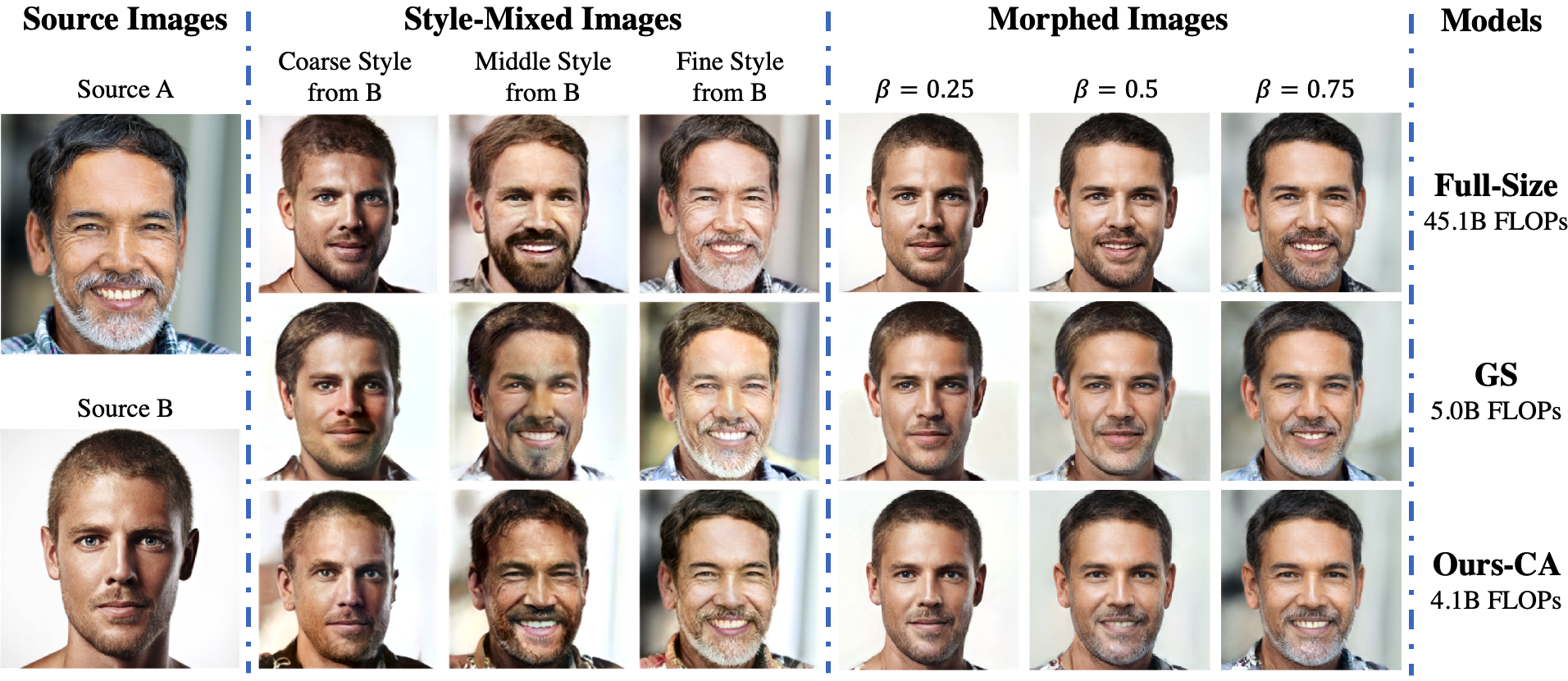}
    \vspace{-0.2cm}
    \caption{An example typifying the effectiveness of our compressed StyleGAN2 for image style-mixing and morphing. 
    When we mix middle styles from B, 
    GS produces a blurred face with uneven skin texture while the original full-size model has a significant identity loss.
    In contrast, our approach better preserves the person's identity with less artifacts.
    We observe that our morphed images have a smoother expression transition compared to GS's in mouth and teeth,
    and our beard transition is even better than the full-size model, 
    substantiating our advantage in latent space smoothness.
    }
    \label{fig:style_mixing_morphing}
    \vspace{-0.4cm}
\end{figure*}
On SN-GAN,
both of our compressed models have no IS loss at 2$\times$ acceleration
with around 0.35 IS gain over \textbf{GS}, as shown in Tab.~\ref{tab:SN_GAN_SOTA}.
At 4$\times$ acceleration level, our method is even more promising:
while \textbf{Ours} has a 0.56 IS increase compared to \textbf{GS}, \textbf{Ours-CA} can further improve \textbf{Ours} by 0.1 on IS.
Such results clearly show the advantage of our content-aware GAN compression scheme.

On 256px StyleGAN2, \textbf{Ours} enjoys an 11$\times$ acceleration from the full-size model 
and achieves 3.5 FID gain over
the $9\times$-accelerated \textbf{GS} model, shown in Tab.~\ref{tab:StyleGAN2_SOTA}. 
\textbf{Ours-CA} further improves \textbf{Ours} by 1.0 on FID and advances the projection performance, 
especially for the $COI$ projection measured by CA-PSNR/CA-LPIPS.
We also note \textbf {Ours-CA} achieves a much lower PPL compared to \textbf{GS}, 
and even smaller than that of the full-size model.
This indicates that our content-aware GAN compression scheme can not only improve model's efficiency, but also the  smoothness of its latent space.
This PPL improvement is further examplified by visual evidence in Fig.~\ref{fig:style_mixing_morphing} and~\ref{fig:ganspace}.

At 1024px resolution, we can only obtain a 3.1$\times$-accelerated generator by \textbf{GS} where overpruning 
would not guarantee the generator to converge in the learning process.
With 3.4$\times$ acceleration over \textbf{GS}, \textbf{Ours}/\textbf{Ours-CA} enjoy an FID improvement of 2.0/2.5 and better image projection performance.
\textbf{Ours-CA} again achieve the best performance in image generation and image projection.





\subsection{Image Editing}

We further demonstrate the benefit of our content-aware compressed StyleGAN2 for editing tasks of style mixing, latent space image morphing, and a recent proposed technique, GANSpace~\cite{harkonen2020ganspace}.
More results are included in the Supplementary.

\textbf{Style Mixing and Morphing.} 
Given two real world images, we run our image projection algorithm in Sec.~\ref{sec:image_projection} to find their $W^+$ embedding latent codes,
followed by two manipulations of these codes~\cite{karras2019style,abdal2019image2stylegan}:
(1) crossover the codes at layer $l \in [1\text{:}L]$ for 
image style-mixing;
(2) interpolate the codes with parameter $\beta \in [0,1]$ for image morphing.

We show an example in Fig.~\ref{fig:style_mixing_morphing} with a 256px model where we set $l\in\{4,8,11\}$ and $\beta\in\{0.25,0.5,0.75\}$. 
While GS yields a number of artifacts (skin texture, hair, etc.) for style-mixing, 
our model performs comparatively to the full-size model in image quality and even preserves a better identity (in middle style mixing).
We find visual evidence for our advantage of having a smoother latent space in the morphed images,
where our expression transition is much smoother than GS in mouth and teeth and 
our beard transition is even better than the full-size model.
This agrees well with the fact that our model have a lower PPL score.

\textbf{GANSpace Editing.}
We further deploy our compressed 1024px model for GANSpace~\cite{harkonen2020ganspace} editing.
Specifically, we use PCA to find principle components of the $W$ space, and traverse a latent code in the direction of a component to generate a sequence of images. 

\begin{figure}[t]
    \centering
    \includegraphics[width=0.48\textwidth]{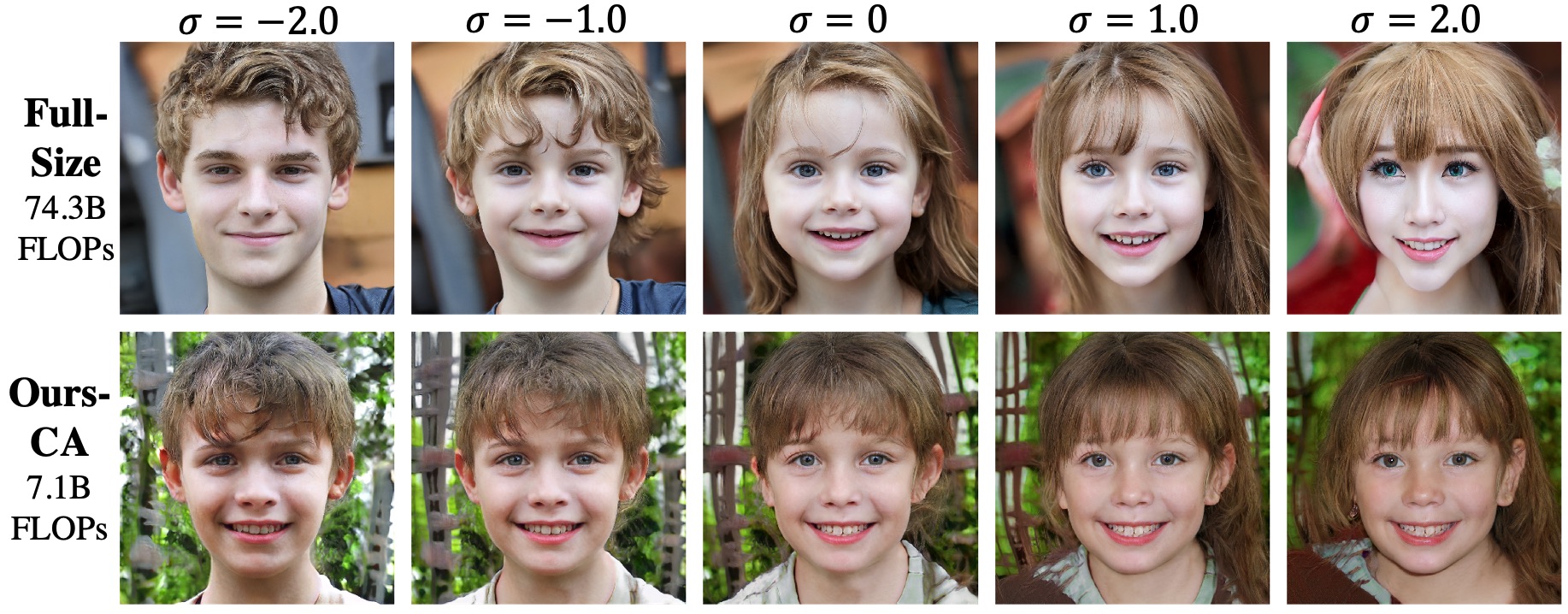}
    \vspace{-0.7cm}
    \caption{A demonstration of more effective GANSpace direction discovery with our compressed model. 
    The direction is suggested for change of gender~\cite{harkonen2020ganspace}, 
    yet the full-size model also changes the age significantly along the direction. 
    In contrast, our compressed model retains the age and produces less artifact at large variation ($\sigma = 2.0$).
    These results suggest that our GANSpace direction is more disentangled, which indicates our latent space is more linearly separable.
    }
    \label{fig:ganspace}
    \vspace{-0.3cm}
\end{figure}

We show an example in  in Fig.~\ref{fig:ganspace}, where we use the same latent code as in the original paper~\cite{harkonen2020ganspace} and traverse it in the direction of the first principal component, $\mathbf{u}_0$.
While it is claimed that $\mathbf{u}_0$ is a direction for gender editing, 
we find that the full-size model also changes person's age along $\mathbf{u}_0$. 
The full-size model also produces artifacts at the chin of the generated images at large deviation.
In contrast, our compressed model only changes the person's gender and generates more natural images at different variation scales. 
This again visually indicates that our compressed model has a smoother and more disentangled latent manifold.

\section{Ablation Study}

\textbf{Pruning Effectiveness.}
We conduct a channel selection analysis on StyleGAN2 with 5 pruning metrics:
low-act~\cite{hu2016network}, $\ell1$-in~\cite{li2020gan}, random, $\ell1$-out, and CA-$\ell1$-out.
Specifically, we create 20 pruned models 
by pruning the original full-size generators with these 5 metrics at 4 layer remove ratios (10\%, 20\%, 30\%, 40\%) without fine-tuning. 

\begin{figure}[t]
    \centering
    \includegraphics[width=0.48\textwidth]{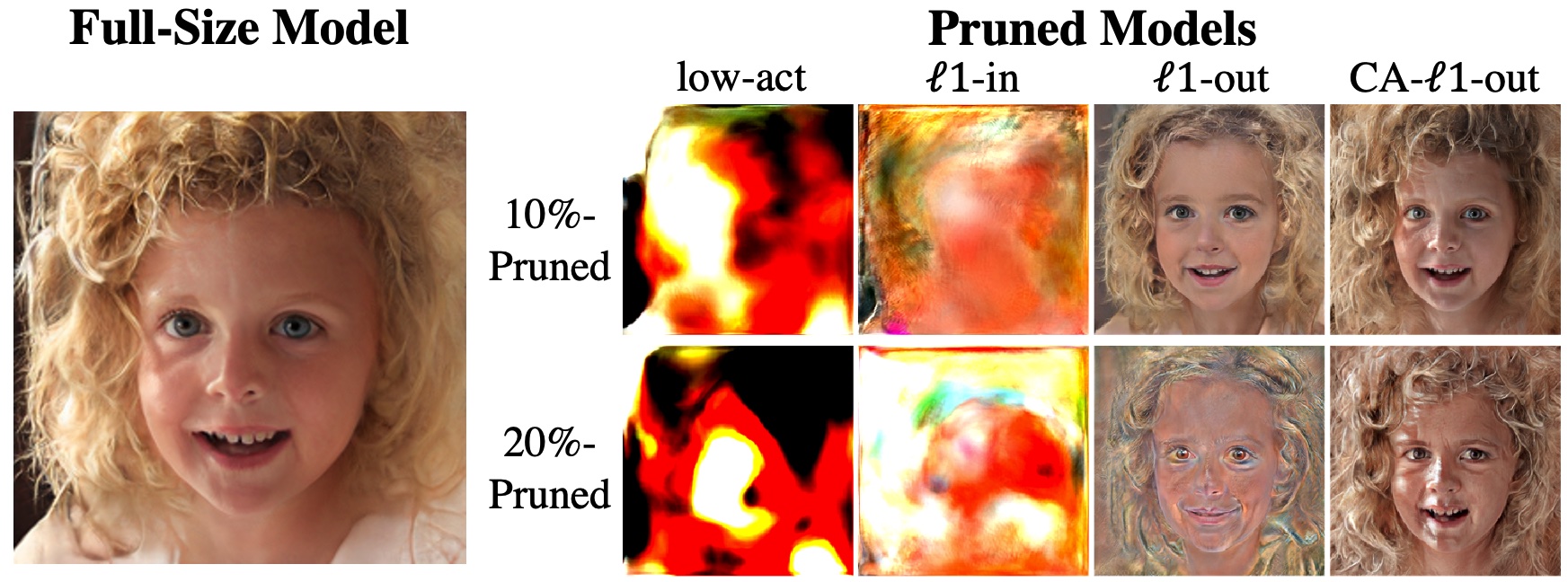}
    \vspace{-0.7cm}
    \caption{Effectiveness of our pruning metrics.
    \textbf{Left:} An image generated by full-size model. \textbf{Right:} Images generated by 8 pruned models (without retraining) varying layer pruning ratios (rows) and pruning metrics (columns).
    We find that both of our metrics $\ell1$-out and CA-$\ell1$-out achieves much better visual quality than low-act~\cite{hu2016network} and $\ell1$-in~\cite{li2020gan}, 
    while CA-$\ell1$-out enjoys the best perceptual performance.
    }
    \label{fig:metric_effectiveness}
\end{figure}

\begin{figure}[t]
    \centering
    \includegraphics[width=0.4\textwidth]{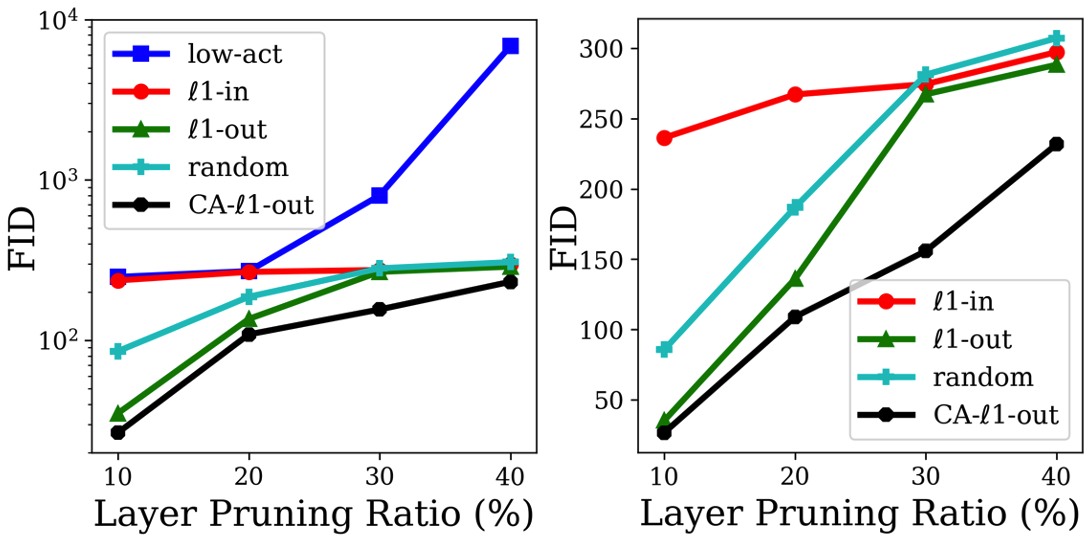}
    \vspace{-0.2cm}
    \caption{FID of pruned models obtained by different pruning metrics at different pruning ratios. \textbf{Left:} FID in log scale with low-act~\cite{hu2016network}. \textbf{Right:} FID in normal scale without low-act. 
    Our CA-$\ell1$-out metric best identifies informative channels quantitatively.}
    \label{fig:metric_FID}
    \vspace{-0.3cm}
\end{figure}

We randomly sample a latent variable and obtain the output images from the full-size model and the pruned models as shown in Fig.~\ref{fig:metric_effectiveness}.
We can clearly find that pruning low-activation (Col.~1) channels would distort the output images completely,
and the prior conditional GAN compression metric, $\ell1$-in (Col.~2), also fails to identify informative channels. 
We find that CA-$\ell1$-out (Col.~4) preserves the most informative channels for image generation. 
We also plot the FID scores of these pruned models in Fig.~\ref{fig:metric_FID},
which correlates well with our visual judgement that CA-$\ell1$-out achieves the best performance.

\begin{figure}[t]
    \centering \includegraphics[width=0.47\textwidth]{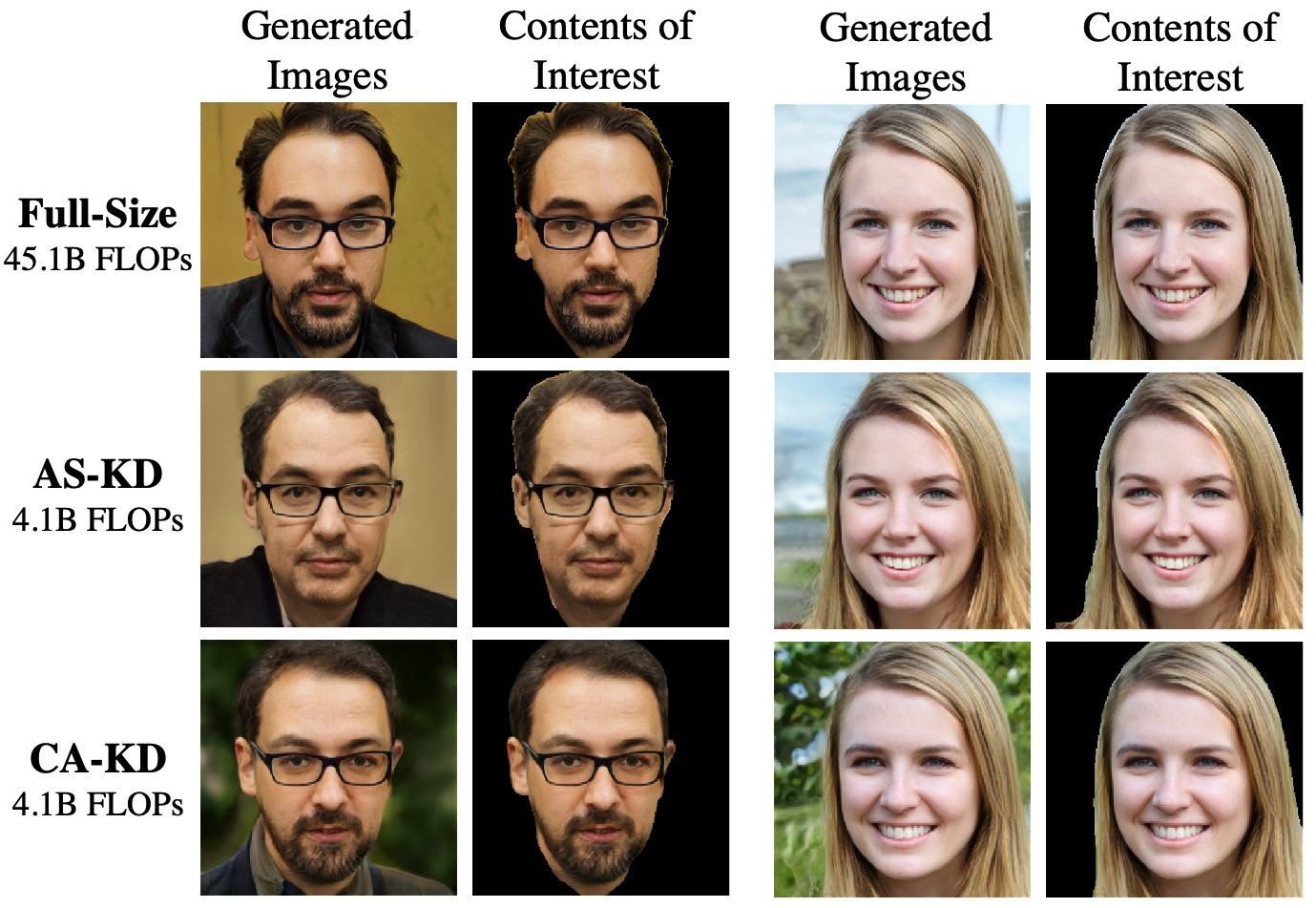}
    \vspace{-0.25cm}
    \caption{ Effectiveness of content-aware distillation (CA-KD) scheme.
    Compared to all spatial locations distillation (AS-KD),
    the model learned by the CA-KD has better identity preservation, more similar glasses, beard and eyes in the content of interest region as its full-size teacher.}
    \label{fig:ca_kd_effectiveness}
\end{figure}

\textbf{Distillation Effectiveness.}
We further demonstrate the advantage of our content-aware knowledge distillation (CA-KD) scheme 
over the all spatial locations distillation (AS-KD) method.
As shown in Fig.~\ref{fig:ca_kd_effectiveness}, 
although the generated images from AS-KD might have similar backgrounds and clothes to the full-size model,
our CA-KD scheme generates images with closer $COI$ features (beard, eyes, glasses, etc.) as its teacher.
Such characteristic not only improves model's generation quality (FID), 
but also explains the enhancement in model's image embedding quality (PSNR/LPIPS), 
especially for the $COI$  region (CA-PSNR/LPIPS),
where it owns a much better distillation.

\section{Conclusion}

In this work, 
we propose a novel content-aware compression pipeline to learn efficient GANs.
While prior works mainly focus on conditional GANs compression,
we study a new approach of channel pruning and knowledge distillation under the context of unconditional GANs,
and further introduce a content-aware version for both compression techniques.
We carry out experiments on SN-GAN and StyleGAN2 to show the effectiveness of our scheme,
where we outperform the state-of-the-art method on all tasks.
Moreover, our compressed models not only enjoy a better resource-performance tradeoff compared to the full-size one,
but also owns an extra advantage of smoother latent space manifold by numerical and visual evidences.

\bibliographystyle{Reference_File/ieee_fullname}
\bibliography{Reference_File/refs}

\begin{thebibliography}{10}\itemsep=-1pt

\bibitem{adobe_photoshop}
{\em Adobe Photoshop}.
\newblock \url{https://www.adobe.com/products/photoshop.html}.

\bibitem{abadi2016tensorflow}
Mart{\'\i}n Abadi, Paul Barham, Jianmin Chen, Zhifeng Chen, Andy Davis, Jeffrey
  Dean, Matthieu Devin, Sanjay Ghemawat, Geoffrey Irving, Michael Isard, et~al.
\newblock Tensorflow: A system for large-scale machine learning.
\newblock In {\em 12th $\{$USENIX$\}$ symposium on operating systems design and
  implementation ($\{$OSDI$\}$ 16)}, pages 265--283, 2016.

\bibitem{abdal2019image2stylegan}
Rameen Abdal, Yipeng Qin, and Peter Wonka.
\newblock Image2stylegan: How to embed images into the stylegan latent space?
\newblock In {\em Proceedings of the IEEE international conference on computer
  vision}, pages 4432--4441, 2019.

\bibitem{abdal2020image2stylegan++}
Rameen Abdal, Yipeng Qin, and Peter Wonka.
\newblock Image2stylegan++: How to edit the embedded images?
\newblock In {\em Proceedings of the IEEE/CVF Conference on Computer Vision and
  Pattern Recognition}, pages 8296--8305, 2020.

\bibitem{aguinaldo2019compressing}
Angeline Aguinaldo, Ping-Yeh Chiang, Alex Gain, Ameya Patil, Kolten Pearson,
  and Soheil Feizi.
\newblock Compressing gans using knowledge distillation.
\newblock {\em arXiv preprint arXiv:1902.00159}, 2019.

\bibitem{badrinarayanan2017segnet}
Vijay Badrinarayanan, Alex Kendall, and Roberto Cipolla.
\newblock Segnet: A deep convolutional encoder-decoder architecture for image
  segmentation.
\newblock {\em IEEE transactions on pattern analysis and machine intelligence},
  39(12):2481--2495, 2017.

\bibitem{brock2018large}
Andrew Brock, Jeff Donahue, and Karen Simonyan.
\newblock Large scale gan training for high fidelity natural image synthesis.
\newblock {\em arXiv preprint arXiv:1809.11096}, 2018.

\bibitem{chang2020tinygan}
Ting-Yun Chang and Chi-Jen Lu.
\newblock Tinygan: Distilling biggan for conditional image generation.
\newblock {\em arXiv preprint arXiv:2009.13829}, 2020.

\bibitem{chen2017learning}
Guobin Chen, Wongun Choi, Xiang Yu, Tony Han, and Manmohan Chandraker.
\newblock Learning efficient object detection models with knowledge
  distillation.
\newblock In {\em Advances in Neural Information Processing Systems}, pages
  742--751, 2017.

\bibitem{chen2020distilling}
Hanting Chen, Yunhe Wang, Han Shu, Changyuan Wen, Chunjing Xu, Boxin Shi, Chao
  Xu, and Chang Xu.
\newblock Distilling portable generative adversarial networks for image
  translation.
\newblock {\em arXiv preprint arXiv:2003.03519}, 2020.

\bibitem{chen2017darkrank}
Yuntao Chen, Naiyan Wang, and Zhaoxiang Zhang.
\newblock Darkrank: Accelerating deep metric learning via cross sample
  similarities transfer.
\newblock {\em arXiv preprint arXiv:1707.01220}, 2017.

\bibitem{courbariaux2016binarized}
Matthieu Courbariaux, Itay Hubara, Daniel Soudry, Ran El-Yaniv, and Yoshua
  Bengio.
\newblock Binarized neural networks: Training deep neural networks with weights
  and activations constrained to+ 1 or-1.
\newblock {\em arXiv preprint arXiv:1602.02830}, 2016.

\bibitem{duan2019content}
Yiping Duan, Yaqiang Zhang, Xiaoming Tao, Chaoyi Han, Mai Xu, Cheng Yang, and
  Jianhua Lu.
\newblock Content-aware deep perceptual image compression.
\newblock In {\em 2019 11th International Conference on Wireless Communications
  and Signal Processing (WCSP)}, pages 1--6. IEEE, 2019.

\bibitem{frechet1957distance}
Maurice Fr{\'e}chet.
\newblock Sur la distance de deux lois de probabilit{\'e}.
\newblock {\em COMPTES RENDUS HEBDOMADAIRES DES SEANCES DE L ACADEMIE DES
  SCIENCES}, 244(6):689--692, 1957.

\bibitem{fu2019dual}
Jun Fu, Jing Liu, Haijie Tian, Yong Li, Yongjun Bao, Zhiwei Fang, and Hanqing
  Lu.
\newblock Dual attention network for scene segmentation.
\newblock In {\em Proceedings of the IEEE Conference on Computer Vision and
  Pattern Recognition}, pages 3146--3154, 2019.

\bibitem{goodfellow2014generative}
Ian Goodfellow, Jean Pouget-Abadie, Mehdi Mirza, Bing Xu, David Warde-Farley,
  Sherjil Ozair, Aaron Courville, and Yoshua Bengio.
\newblock Generative adversarial nets.
\newblock In {\em Advances in neural information processing systems}, pages
  2672--2680, 2014.

\bibitem{han2015learning}
Song Han, Jeff Pool, John Tran, and William Dally.
\newblock Learning both weights and connections for efficient neural network.
\newblock In {\em Advances in neural information processing systems}, pages
  1135--1143, 2015.

\bibitem{harkonen2020ganspace}
Erik H{\"a}rk{\"o}nen, Aaron Hertzmann, Jaakko Lehtinen, and Sylvain Paris.
\newblock Ganspace: Discovering interpretable gan controls.
\newblock {\em arXiv preprint arXiv:2004.02546}, 2020.

\bibitem{he2018soft}
Yang He, Guoliang Kang, Xuanyi Dong, Yanwei Fu, and Yi Yang.
\newblock Soft filter pruning for accelerating deep convolutional neural
  networks.
\newblock {\em arXiv preprint arXiv:1808.06866}, 2018.

\bibitem{he2019filter}
Yang He, Ping Liu, Ziwei Wang, Zhilan Hu, and Yi Yang.
\newblock Filter pruning via geometric median for deep convolutional neural
  networks acceleration.
\newblock In {\em Proceedings of the IEEE Conference on Computer Vision and
  Pattern Recognition}, pages 4340--4349, 2019.

\bibitem{heusel2017gans}
Martin Heusel, Hubert Ramsauer, Thomas Unterthiner, Bernhard Nessler, and Sepp
  Hochreiter.
\newblock Gans trained by a two time-scale update rule converge to a local nash
  equilibrium.
\newblock In {\em Advances in neural information processing systems}, pages
  6626--6637, 2017.

\bibitem{hinton2015distilling}
Geoffrey Hinton, Oriol Vinyals, and Jeff Dean.
\newblock Distilling the knowledge in a neural network.
\newblock {\em arXiv preprint arXiv:1503.02531}, 2015.

\bibitem{hou2017deeply}
Qibin Hou, Ming-Ming Cheng, Xiaowei Hu, Ali Borji, Zhuowen Tu, and Philip~HS
  Torr.
\newblock Deeply supervised salient object detection with short connections.
\newblock In {\em Proceedings of the IEEE Conference on Computer Vision and
  Pattern Recognition}, pages 3203--3212, 2017.

\bibitem{hu2016network}
Hengyuan Hu, Rui Peng, Yu-Wing Tai, and Chi-Keung Tang.
\newblock Network trimming: A data-driven neuron pruning approach towards
  efficient deep architectures.
\newblock {\em arXiv preprint arXiv:1607.03250}, 2016.

\bibitem{isola2017image}
Phillip Isola, Jun-Yan Zhu, Tinghui Zhou, and Alexei~A Efros.
\newblock Image-to-image translation with conditional adversarial networks.
\newblock In {\em Proceedings of the IEEE conference on computer vision and
  pattern recognition}, pages 1125--1134, 2017.

\bibitem{jacob2018quantization}
Benoit Jacob, Skirmantas Kligys, Bo Chen, Menglong Zhu, Matthew Tang, Andrew
  Howard, Hartwig Adam, and Dmitry Kalenichenko.
\newblock Quantization and training of neural networks for efficient
  integer-arithmetic-only inference.
\newblock In {\em Proceedings of the IEEE Conference on Computer Vision and
  Pattern Recognition}, pages 2704--2713, 2018.

\bibitem{jaderberg2014speeding}
Max Jaderberg, Andrea Vedaldi, and Andrew Zisserman.
\newblock Speeding up convolutional neural networks with low rank expansions.
\newblock In {\em Proceedings of the British Machine Vision Conference. BMVA
  Press}, 2014.

\bibitem{johnson2016perceptual}
Justin Johnson, Alexandre Alahi, and Li Fei-Fei.
\newblock Perceptual losses for real-time style transfer and super-resolution.
\newblock In {\em European conference on computer vision}, pages 694--711.
  Springer, 2016.

\bibitem{karras2019style}
Tero Karras, Samuli Laine, and Timo Aila.
\newblock A style-based generator architecture for generative adversarial
  networks.
\newblock In {\em Proceedings of the IEEE conference on computer vision and
  pattern recognition}, pages 4401--4410, 2019.

\bibitem{karras2020analyzing}
Tero Karras, Samuli Laine, Miika Aittala, Janne Hellsten, Jaakko Lehtinen, and
  Timo Aila.
\newblock Analyzing and improving the image quality of stylegan.
\newblock In {\em Proceedings of the IEEE/CVF Conference on Computer Vision and
  Pattern Recognition}, pages 8110--8119, 2020.

\bibitem{kingma2014adam}
Diederik~P Kingma and Jimmy Ba.
\newblock Adam: A method for stochastic optimization.
\newblock {\em arXiv preprint arXiv:1412.6980}, 2014.

\bibitem{krizhevsky2009learning}
Alex Krizhevsky, Geoffrey Hinton, et~al.
\newblock Learning multiple layers of features from tiny images.
\newblock 2009.

\bibitem{le2012interactive}
Vuong Le, Jonathan Brandt, Zhe Lin, Lubomir Bourdev, and Thomas~S Huang.
\newblock Interactive facial feature localization.
\newblock In {\em European conference on computer vision}, pages 679--692.
  Springer, 2012.

\bibitem{lebedev2015speeding}
V Lebedev, Y Ganin, M Rakhuba, I Oseledets, and V Lempitsky.
\newblock Speeding-up convolutional neural networks using fine-tuned
  cp-decomposition.
\newblock In {\em 3rd International Conference on Learning Representations,
  ICLR 2015-Conference Track Proceedings}, 2015.

\bibitem{li2016pruning}
Hao Li, Asim Kadav, Igor Durdanovic, Hanan Samet, and Hans~Peter Graf.
\newblock Pruning filters for efficient convnets.
\newblock {\em arXiv preprint arXiv:1608.08710}, 2016.

\bibitem{li2020gan}
Muyang Li, Ji Lin, Yaoyao Ding, Zhijian Liu, Jun-Yan Zhu, and Song Han.
\newblock Gan compression: Efficient architectures for interactive conditional
  gans.
\newblock In {\em Proceedings of the IEEE/CVF Conference on Computer Vision and
  Pattern Recognition}, pages 5284--5294, 2020.

\bibitem{li2020few}
Tianhong Li, Jianguo Li, Zhuang Liu, and Changshui Zhang.
\newblock Few sample knowledge distillation for efficient network compression.
\newblock In {\em Proceedings of the IEEE/CVF Conference on Computer Vision and
  Pattern Recognition}, pages 14639--14647, 2020.

\bibitem{liu1989limited}
Dong~C Liu and Jorge Nocedal.
\newblock On the limited memory bfgs method for large scale optimization.
\newblock {\em Mathematical programming}, 45(1-3):503--528, 1989.

\bibitem{liu2017learning}
Zhuang Liu, Jianguo Li, Zhiqiang Shen, Gao Huang, Shoumeng Yan, and Changshui
  Zhang.
\newblock Learning efficient convolutional networks through network slimming.
\newblock In {\em Proceedings of the IEEE International Conference on Computer
  Vision}, pages 2736--2744, 2017.

\bibitem{miyato2018spectral}
Takeru Miyato, Toshiki Kataoka, Masanori Koyama, and Yuichi Yoshida.
\newblock Spectral normalization for generative adversarial networks.
\newblock {\em arXiv preprint arXiv:1802.05957}, 2018.

\bibitem{rastegari2016xnor}
Mohammad Rastegari, Vicente Ordonez, Joseph Redmon, and Ali Farhadi.
\newblock Xnor-net: Imagenet classification using binary convolutional neural
  networks.
\newblock In {\em European conference on computer vision}, pages 525--542.
  Springer, 2016.

\bibitem{romero2014fitnets}
Adriana Romero, Nicolas Ballas, Samira~Ebrahimi Kahou, Antoine Chassang, Carlo
  Gatta, and Yoshua Bengio.
\newblock Fitnets: Hints for thin deep nets.
\newblock {\em arXiv preprint arXiv:1412.6550}, 2014.

\bibitem{salimans2016improved}
Tim Salimans, Ian Goodfellow, Wojciech Zaremba, Vicki Cheung, Alec Radford, and
  Xi Chen.
\newblock Improved techniques for training gans.
\newblock In {\em Advances in neural information processing systems}, pages
  2234--2242, 2016.

\bibitem{sandler2018mobilenetv2}
Mark Sandler, Andrew Howard, Menglong Zhu, Andrey Zhmoginov, and Liang-Chieh
  Chen.
\newblock Mobilenetv2: Inverted residuals and linear bottlenecks.
\newblock In {\em Proceedings of the IEEE conference on computer vision and
  pattern recognition}, pages 4510--4520, 2018.

\bibitem{selvaraju2017grad}
Ramprasaath~R Selvaraju, Michael Cogswell, Abhishek Das, Ramakrishna Vedantam,
  Devi Parikh, and Dhruv Batra.
\newblock Grad-cam: Visual explanations from deep networks via gradient-based
  localization.
\newblock In {\em Proceedings of the IEEE international conference on computer
  vision}, pages 618--626, 2017.

\bibitem{shu2019co}
Han Shu, Yunhe Wang, Xu Jia, Kai Han, Hanting Chen, Chunjing Xu, Qi Tian, and
  Chang Xu.
\newblock Co-evolutionary compression for unpaired image translation.
\newblock In {\em Proceedings of the IEEE International Conference on Computer
  Vision}, pages 3235--3244, 2019.

\bibitem{szegedy2016rethinking}
Christian Szegedy, Vincent Vanhoucke, Sergey Ioffe, Jon Shlens, and Zbigniew
  Wojna.
\newblock Rethinking the inception architecture for computer vision.
\newblock In {\em Proceedings of the IEEE conference on computer vision and
  pattern recognition}, pages 2818--2826, 2016.

\bibitem{viazovetskyi2020stylegan2}
Yuri Viazovetskyi, Vladimir Ivashkin, and Evgeny Kashin.
\newblock Stylegan2 distillation for feed-forward image manipulation.
\newblock {\em arXiv preprint arXiv:2003.03581}, 2020.

\bibitem{wang2020gan}
Haotao Wang, Shupeng Gui, Haichuan Yang, Ji Liu, and Zhangyang Wang.
\newblock Gan slimming: All-in-one gan compression by a unified optimization
  framework.
\newblock {\em arXiv preprint arXiv:2008.11062}, 2020.

\bibitem{woo2018cbam}
Sanghyun Woo, Jongchan Park, Joon-Young Lee, and In So~Kweon.
\newblock Cbam: Convolutional block attention module.
\newblock In {\em Proceedings of the European conference on computer vision
  (ECCV)}, pages 3--19, 2018.

\bibitem{ye2018rethinking}
Jianbo Ye, Xin Lu, Zhe Lin, and James~Z Wang.
\newblock Rethinking the smaller-norm-less-informative assumption in channel
  pruning of convolution layers.
\newblock {\em arXiv preprint arXiv:1802.00124}, 2018.

\bibitem{yim2017gift}
Junho Yim, Donggyu Joo, Jihoon Bae, and Junmo Kim.
\newblock A gift from knowledge distillation: Fast optimization, network
  minimization and transfer learning.
\newblock In {\em Proceedings of the IEEE Conference on Computer Vision and
  Pattern Recognition}, pages 4133--4141, 2017.

\bibitem{yu2018bisenet}
Changqian Yu, Jingbo Wang, Chao Peng, Changxin Gao, Gang Yu, and Nong Sang.
\newblock Bisenet: Bilateral segmentation network for real-time semantic
  segmentation.
\newblock In {\em Proceedings of the European conference on computer vision
  (ECCV)}, pages 325--341, 2018.

\bibitem{zagoruyko2016paying}
Sergey Zagoruyko and Nikos Komodakis.
\newblock Paying more attention to attention: Improving the performance of
  convolutional neural networks via attention transfer.
\newblock {\em arXiv preprint arXiv:1612.03928}, 2016.

\bibitem{zhang2018unreasonable}
Richard Zhang, Phillip Isola, Alexei~A Efros, Eli Shechtman, and Oliver Wang.
\newblock The unreasonable effectiveness of deep features as a perceptual
  metric.
\newblock In {\em Proceedings of the IEEE conference on computer vision and
  pattern recognition}, pages 586--595, 2018.

\bibitem{zhang2018systematic}
Tianyun Zhang, Shaokai Ye, Kaiqi Zhang, Jian Tang, Wujie Wen, Makan Fardad, and
  Yanzhi Wang.
\newblock A systematic dnn weight pruning framework using alternating direction
  method of multipliers.
\newblock In {\em Proceedings of the European Conference on Computer Vision
  (ECCV)}, pages 184--199, 2018.

\bibitem{zhao2019pyramid}
Ting Zhao and Xiangqian Wu.
\newblock Pyramid feature attention network for saliency detection.
\newblock In {\em Proceedings of the IEEE Conference on Computer Vision and
  Pattern Recognition}, pages 3085--3094, 2019.

\bibitem{zhou2016learning}
Bolei Zhou, Aditya Khosla, Agata Lapedriza, Aude Oliva, and Antonio Torralba.
\newblock Learning deep features for discriminative localization.
\newblock In {\em Proceedings of the IEEE conference on computer vision and
  pattern recognition}, pages 2921--2929, 2016.

\bibitem{zhu2017unpaired}
Jun-Yan Zhu, Taesung Park, Phillip Isola, and Alexei~A Efros.
\newblock Unpaired image-to-image translation using cycle-consistent
  adversarial networks.
\newblock In {\em Proceedings of the IEEE international conference on computer
  vision}, pages 2223--2232, 2017.

\bibitem{zund2013content}
Fabio Z{\"u}nd, Yael Pritch, Alexander Sorkine-Hornung, Stefan Mangold, and
  Thomas Gross.
\newblock Content-aware compression using saliency-driven image retargeting.
\newblock In {\em 2013 IEEE International Conference on Image Processing},
  pages 1845--1849. IEEE, 2013.

\end{thebibliography}
\clearpage

\section{Supplementary Material}

We organize our supplementary material as follows.
In Sec.~\ref{sec:experiment}, we include detailed settings of the experiments.
We discuss additional ablation studies in Sec.~\ref{sec:ablation}. 
We present more generated images from the SN-GAN, 256px StyleGAN2, and 1024px StyleGAN2 in Sec.~\ref{sec:image_generation}.
We show more edited images by latent space image morphing and GANSpace~\cite{harkonen2020ganspace} image editing in Sec.~\ref{sec:image_editing}.

\subsection{Implementation Details}\label{sec:experiment}

\textbf{SN-GAN.} We adopt the public implementation of SN-GAN\footnote{https://github.com/GongXinyuu/sngan.pytorch}. 
We produce the pruned generators by uniformly removing 30\% and 50\% of channels from the full-size generator with our pruning metrics.
We then retrain the pruned models using the Adam~\cite{kingma2014adam} optimizer.
The generator and discriminator are fine-tuned by 320 epochs with a learning rate of 0.0002.
The batch sizes for generator and discriminator are 128 and 64, respectively.
We update 1 step of the discriminator after 5 generator iterations.

\textbf{StyleGAN2.} We adopt the public implementation of StyleGAN2\footnote{https://github.com/rosinality/stylegan2-pytorch}.
We uniformly remove 70\% of channels from the full-size generator to produce our pruned models.
We fine-tune the pruned models using the Adam optimizer with 470,000 iterations and a batch size of 16.
For perceptual loss distillation on 1024px StyleGAN2, 
we resize the outputs from both the teacher generator and the student generator to 256px to avoid running out of the memory.
Other training settings are the same as in~\cite{karras2020analyzing}.

We implement the GAN-Slimming~\cite{wang2020gan} on StyleGAN2 by imposing a $\ell1$ sparsity loss on the style scalars which are used for weight modulation.
The knowledge distillation is achieved by the VGG style transfer loss~\cite{johnson2016perceptual}.
Other training settings are the same as our method.

\begin{figure*}
    \centering
    \includegraphics[width=\textwidth]{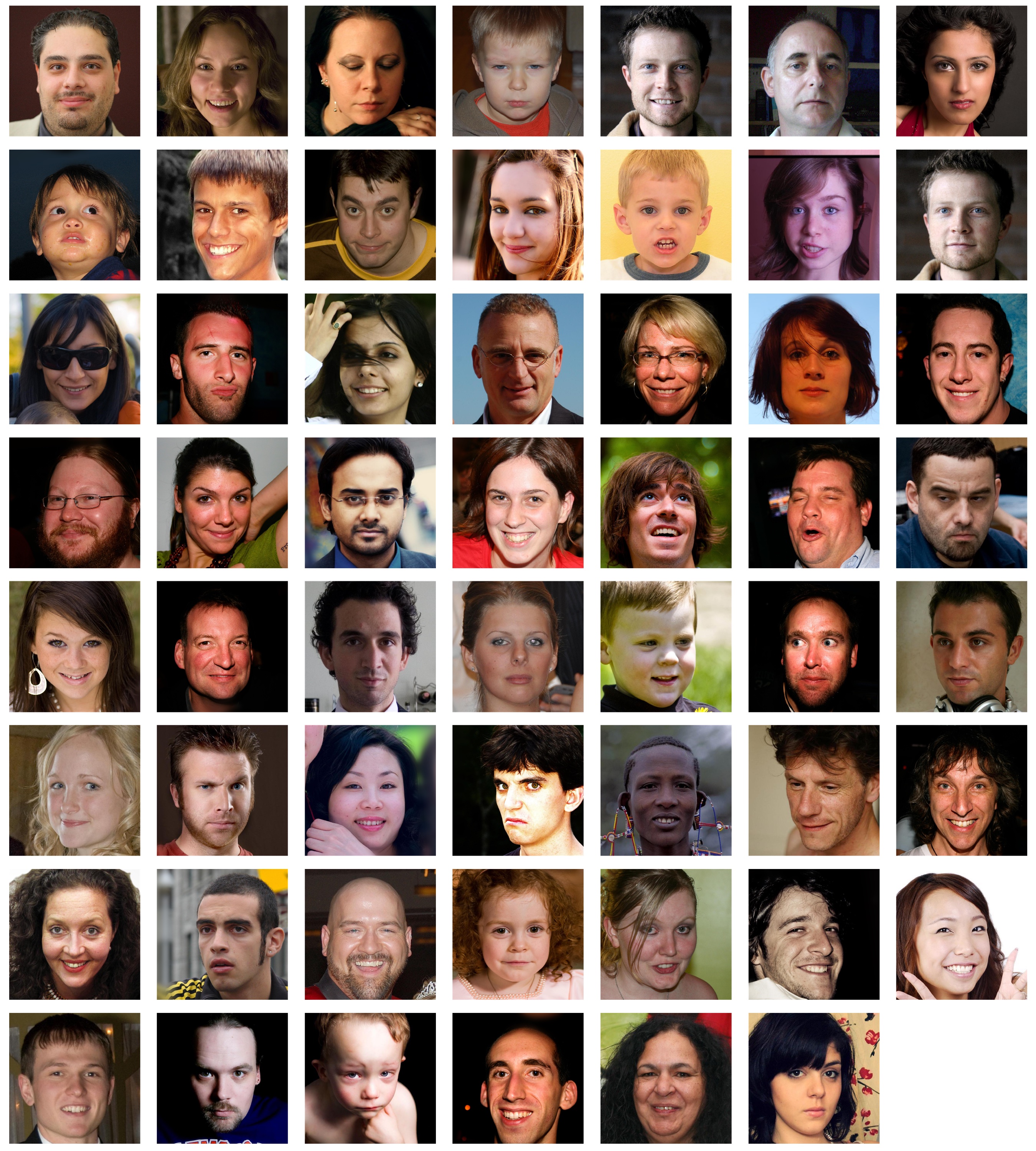}
    \caption{Our dataset, Helen-Set55, with real world images for image projection evaluation.
    This set of images have various attributes of lighting conditions, genders, ethnicities, and ages, 
    which allows a comprehensive evaluation of a model's image projection ability.}
    \label{fig:helen_set_55}
\end{figure*}

\textbf{Image Projection.} 
We present the Helen-Set55 dataset in Fig.~\ref{fig:helen_set_55}, which is adopted for our image projection evaluation on StyleGAN2.
This set of images is sampled from a dataset (Helen~\cite{le2012interactive}) of real world images 
with various lighting conditions, genders, ethnicities, ages, etc.
Such diversity allows a comprehensive evaluation on the embedding performance of the generator.

\subsection{Additional Ablation Study}\label{sec:ablation}

\begin{figure}
    \centering
    \includegraphics[width = 0.45\textwidth]{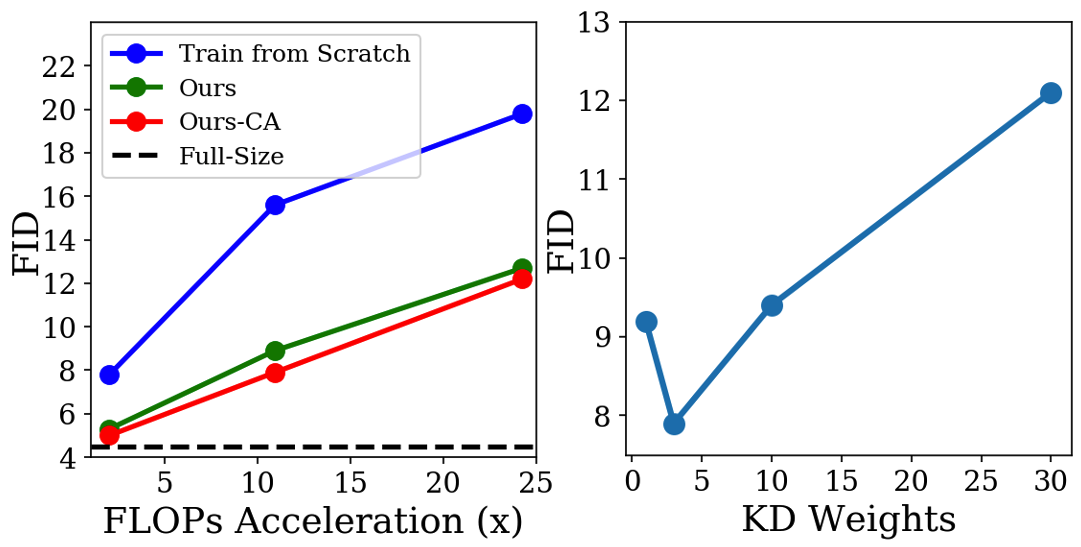}
    \caption{\textbf{Left:} The FID curves of different compression schemes varying acceleration ratio. 
    Content-aware scheme achieves the best performance.
    \textbf{Right:} The FID curve for content-aware compressed 11$\times$-accelerated models varying distillation weights. 
    $\lambda=\gamma=3$ achieves the best results.
    }
    \label{fig:abaltion_hyperparams}
\end{figure}

\textbf{Acceleration Ratio.}
We change the ratio of FLOPs acceleration on 256px StyleGAN2 to comprehensively compare the complexity-performance trade-off.
The result is shown in Fig.~\ref{fig:abaltion_hyperparams},
where we study our scheme at three FLOPs acceleration ratios: 2$\times$, 11$\times$, and 24$\times$.
We find that all of our schemes consistently outperform the training from scratch baseline with a clear margin at each ratio.
The content-aware scheme again achieves the best performance.

\textbf{Knowledge Distillation Weights.}
We investigate into the knowledge distillation weights for our content-aware distillation.
We set $\lambda=\gamma$ and vary them to be $\{1, 3, 10, 30\}$ and test on a 11$\times$-accelerated 256px StyleGAN2. 
We plot FID vs. KD weights in Fig.~\ref{fig:abaltion_hyperparams}.
We find that simply increasing the KD weights would not benefit the student's generation ability.
Moreover, we find that $\lambda=\gamma=3$ achieves the best result out of these four choices.

\begin{table*}[t]

\fontsize{8.5}{10}\selectfont

\centering

\begin{tabular}{|c|c||c|c||c|c|c|}
\hline
Model & Img. Size  &  FID  & Gen. Time Per Img. & PSNR & LPIPS & Proj. Time Per Img.\\
\hline\hline
 
 Original Full-Size & 256 & 4.5 & 11.09ms & 32.02 & 0.113 & 7.57s \\
\hline
\multicolumn{7}{|c|}{Compressed Models} \\
\hline

GS~\cite{wang2020gan} (Our Impl.) & 256  & 12.4 & 2.81ms & 31.02 & 0.177 & 2.51s \\

Ours-CA  & 256 &  \textbf{7.9}  & \textbf{2.67ms} & \textbf{31.41} & \textbf{0.144} &  \textbf{2.44s} \\

\hline
\multicolumn{7}{c}{}\\
\hline

 Original Full-Size & 1024 & 2.7 & 22.46ms & 31.38 & 0.149 & 13.10s \\

\hline
\multicolumn{7}{|c|}{Compressed Models} \\
\hline

GS~\cite{wang2020gan} (Our Impl.) & 1024  &  10.1 & 12.75ms & 30.74 & 0.189 & 9.25s \\

Ours-CA  & 1024 &  \textbf{7.6} & \textbf{5.50ms} & \textbf{30.96} & \textbf{0.170} & \textbf{4.74s} \\

\hline
\end{tabular}

\vspace{-0.3cm}
\caption{Hardware inference acceleration measured with 256px and 1024px StyleGAN2 on a V100 GPU. Our model has a speedup from the full-size model by 4$\times$ on image generation and 3$\times$ on image projection.}
\label{tab:StyleGAN2_SOTA_run_time}
\vspace{-0.4cm}
\end{table*}

\begin{figure}
    \centering
    \includegraphics[width = 0.47\textwidth]{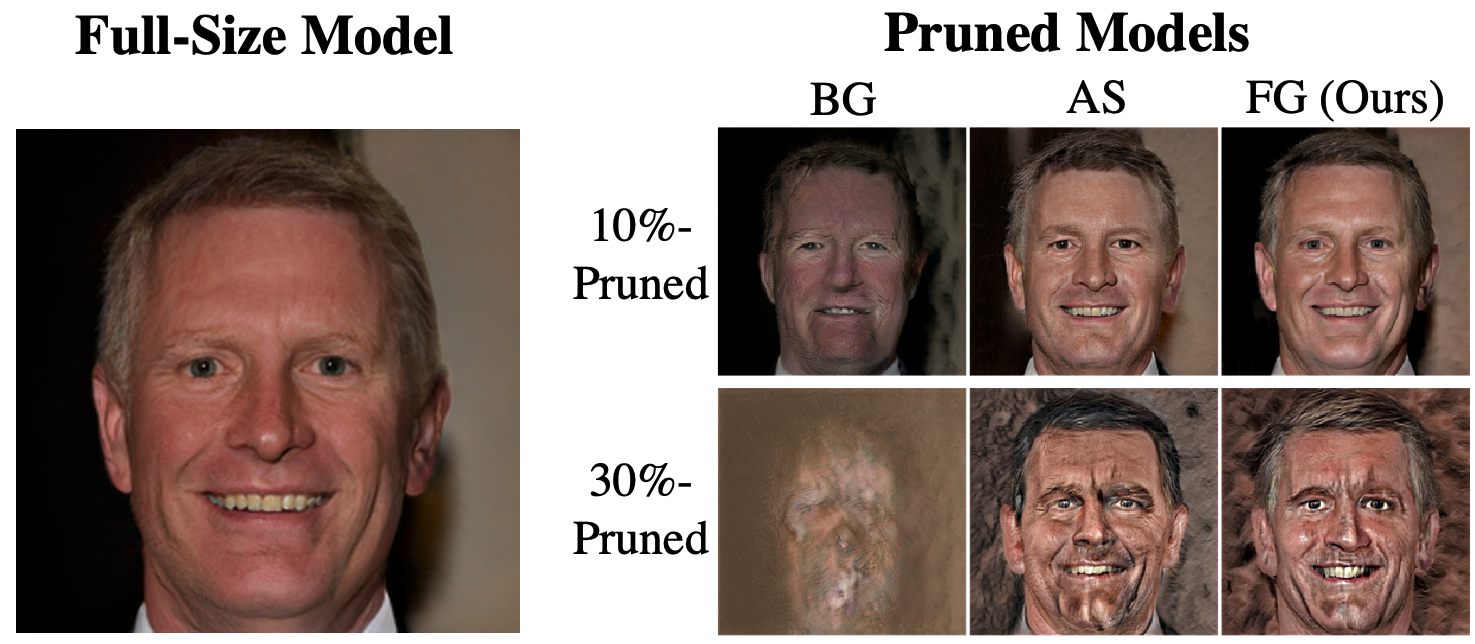}
    \caption{The choice of content of interest ($COI$) in designing the content-aware pruning metric.
    We observe that setting background as $COI$ for pruning (Col.~1) will destroy the useful information from the original generator completely.
    Setting the foreground as $COI$ (Col.~3) demonstrates a clear improvement over setting all spatial locations as $COI$ (Col.~2), especially under a high pruning ratio.
    }
    \label{fig:COI_choice}
\end{figure}

\textbf{COI of Content-Aware Pruning.} 
We inspect the choice of $COI$ in designing our content-aware pruning metric in Fig.~\ref{fig:COI_choice}.
In particular, we show qualitative images from the pruned networks obtained by setting the $COI$ as: the foreground human faces as adopted in the main paper (FG), 
all the spatial locations (AS), and the image background (BG).
We follow the same pruning metric visual analysis as in the main paper's ablation study.
We find that setting the $COI$ as the image background (Col.~1) will completely distort the interested information in the pruning process which is undesirable.
Although AS (Col.~2) performs comparatively compared to our FG (Col.~3) approach at a low pruning ratio (10\%),  
FG retains much more information on the person's faces (teeth, eyes, hair, chin, etc.)
at a higher remove ratio (30\%), 
demonstrating the advantage of choosing the foreground of the image as $COI$.

\textbf{Hardware Inference Speedup.} 
To verify the practical effectiveness of the compressed models, 
we measure the run-time of image generation and image projection\footnote{We run 200 optimization steps for to project an image.} with an Nvidia V100 GPU,
as shown in Tab.~\ref{tab:StyleGAN2_SOTA_run_time}.
Specifically, we measure the time by feed-forwarding/projecting images in a batch manner. 
Due to the variation of memory consumption,
the suitable  batch sizes are different for different models.
For example, 
the largest batch sizes for image generation with full-size 256px and 1024px StyleGAN2 are 16 and 64, respectively. 
Therefore, we divide the generation/projection time over the batch size for a per image based run-time comparison.
Our models can generate/project images faster than GS with better generation/projection quality.
Notably, our models have around 4$\times$
and $3\times$ speedups from the full-size models for image generation and projection, respectively.

\subsection{Additional Image Generation Results}\label{sec:image_generation}

\begin{figure}[t]
    \centering
    \includegraphics[width=0.45\textwidth]{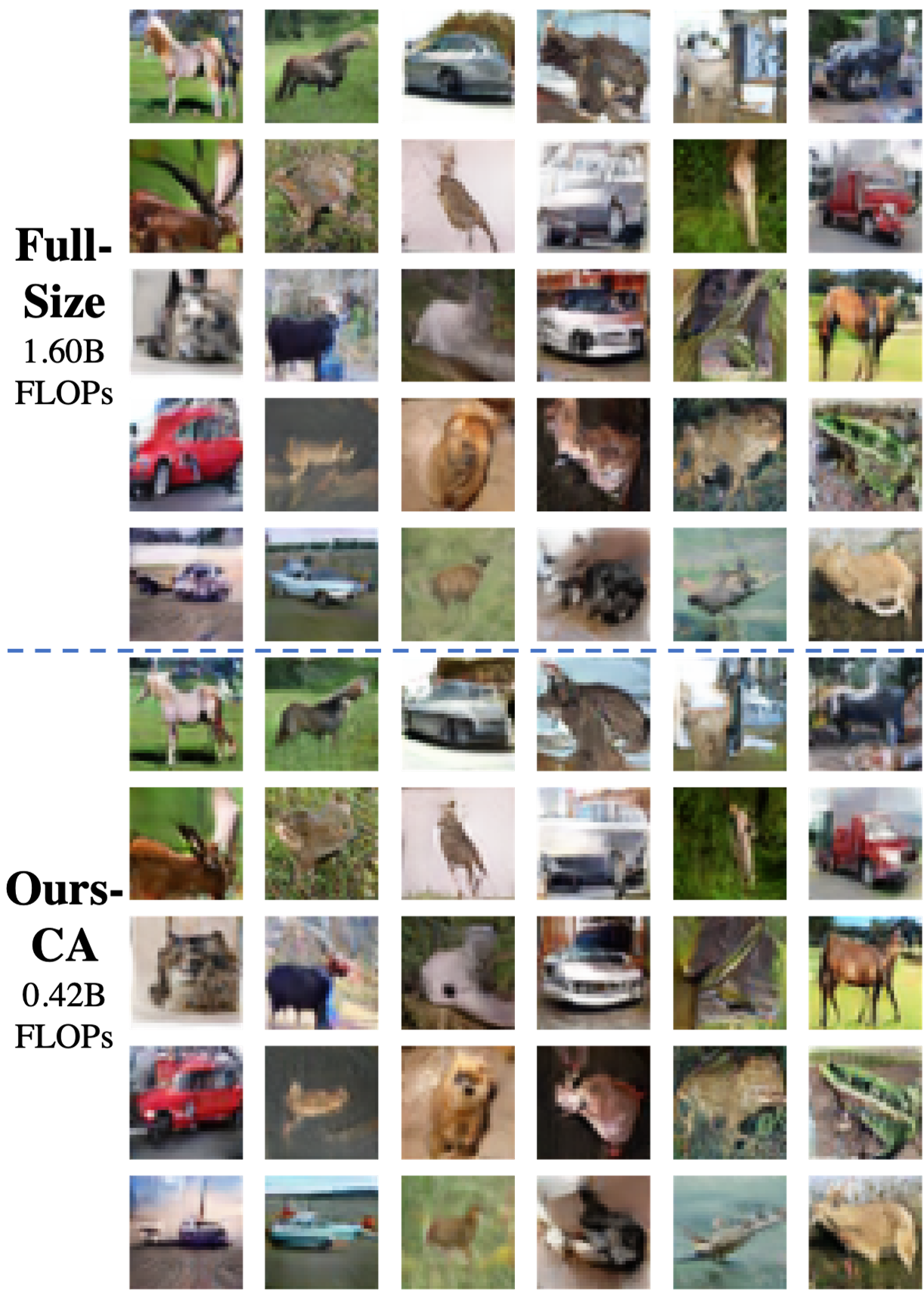}
    \vspace{-0.3cm}
    \caption{Generated images from SN-GAN. 
    Our compressed model generates images with the same quality as the full-size model. }
    \label{fig:sn-gan}
\end{figure}

\begin{figure*}
    \centering
    \includegraphics[width=0.75\textwidth]{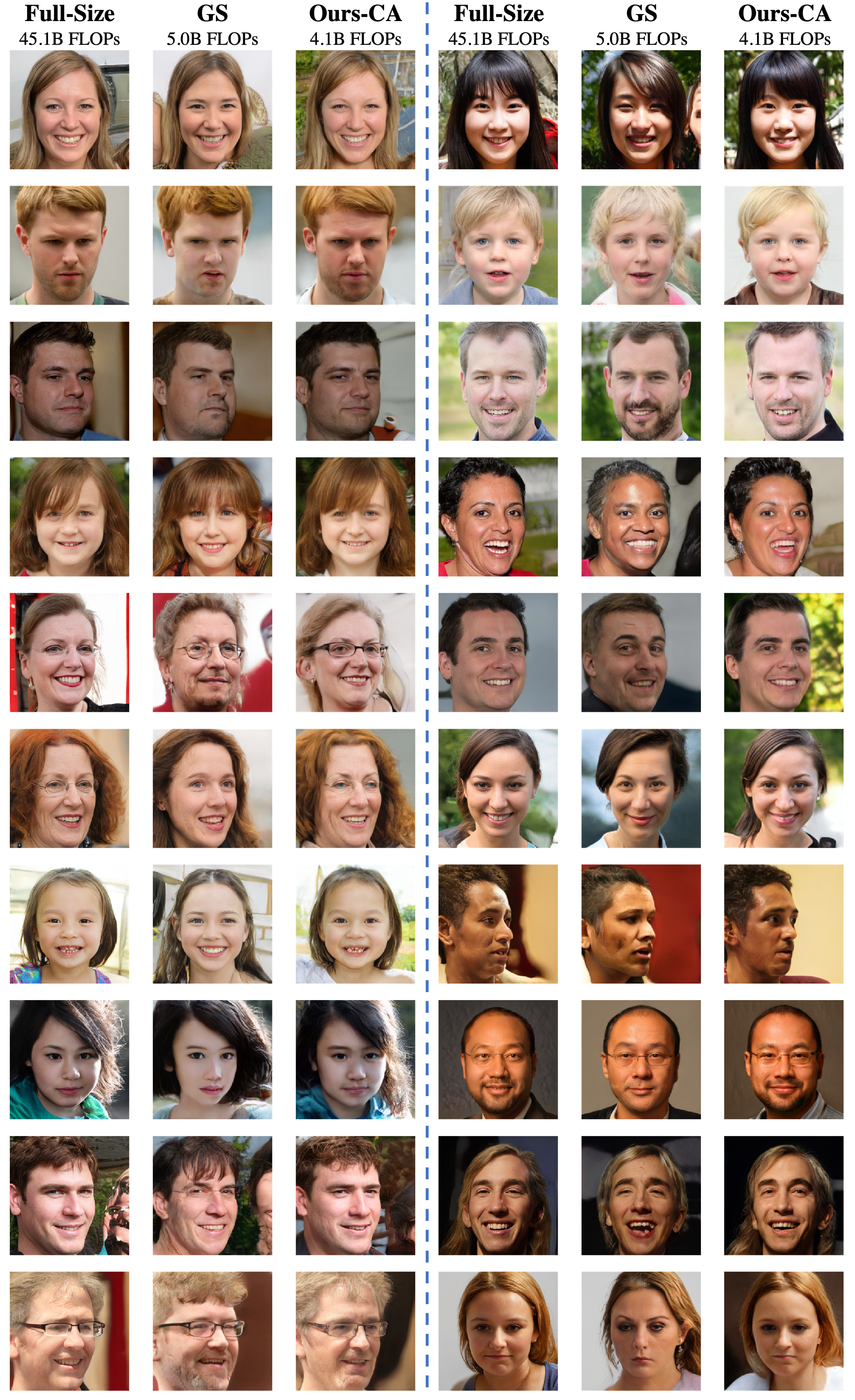}
    \vspace{-0.3cm}
    \caption{Additional generated images from 256px StyleGAN2. 
    Our model shows a better distillation quality and a better image generation quality with a higher acceleration ratio compared to GS.}
    \label{fig:256px_StyleGAN2}
\end{figure*}

\begin{figure*}
    \centering    \includegraphics[width=0.97\textwidth, clip]{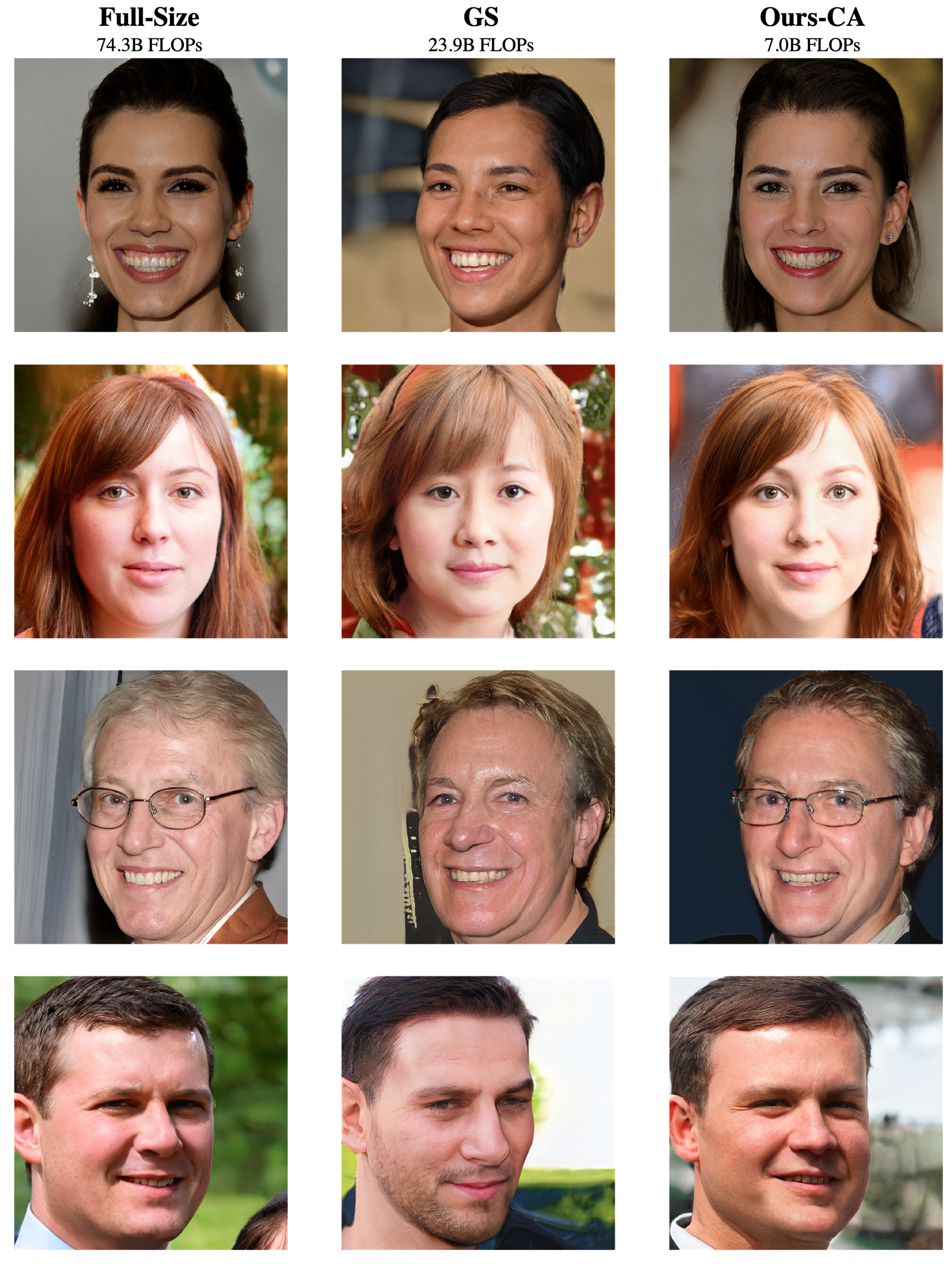}
    \vspace{-0.5cm}
    \caption{Additional generated images from 1024px StyleGAN2. 
    Our model shows a better distillation quality and a better image generation quality with a higher acceleration ratio compared to GS.}
    \label{fig:1024px_StyleGAN2}
\end{figure*}

We show additional SN-GAN generation results in Fig.~\ref{fig:sn-gan} and 256px/1024px StyleGAN2 results in Fig.~\ref{fig:256px_StyleGAN2} and~\ref{fig:1024px_StyleGAN2}. 
Our compressed models generate images with negligible visual quality loss compared to the original full-size model.

\subsection{Additional Image Editing Results}\label{sec:image_editing}

\begin{figure*}[t]
    \centering
    \includegraphics[width=0.98\textwidth]{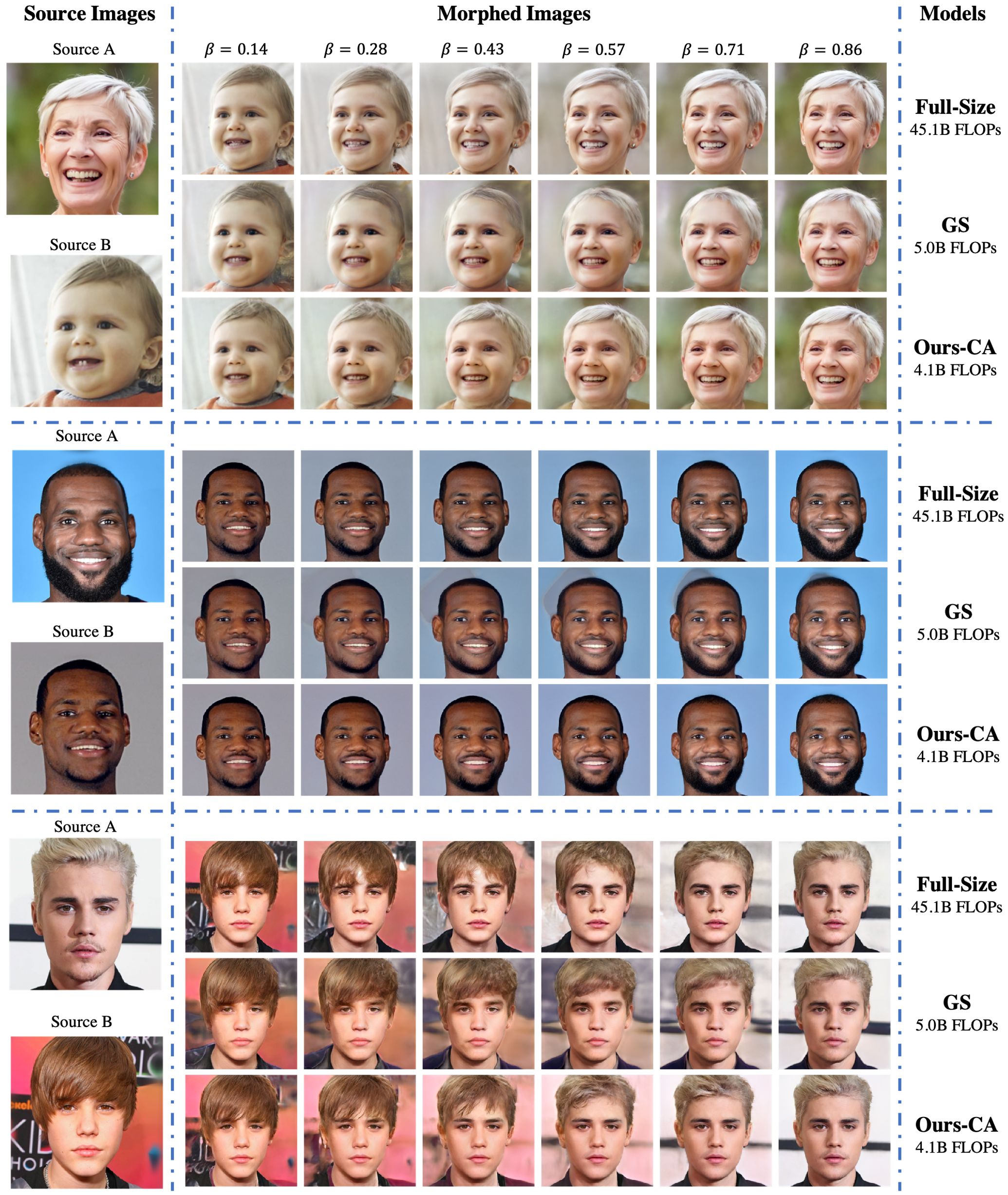}
    \vspace{-0.3cm}
    \caption{Additional examples of latent space image morphing.
    While GS~\cite{wang2020gan} produces a number of artifacts in the morphed images (eyes in the first example, background around the hair in the second examples, etc.),
    our results are of comparative quality as the full-size model. 
    Moreover, we again fine a smoother transition (ages, expressions, eyes, etc.) in our compressed model compared to the full-size one.
    These results substantiate our advantage in latent manifold smoothness.}
    \label{fig:image_morphing}
\end{figure*}

\begin{figure*}[t]
    \centering
    \includegraphics[width=0.87\textwidth]{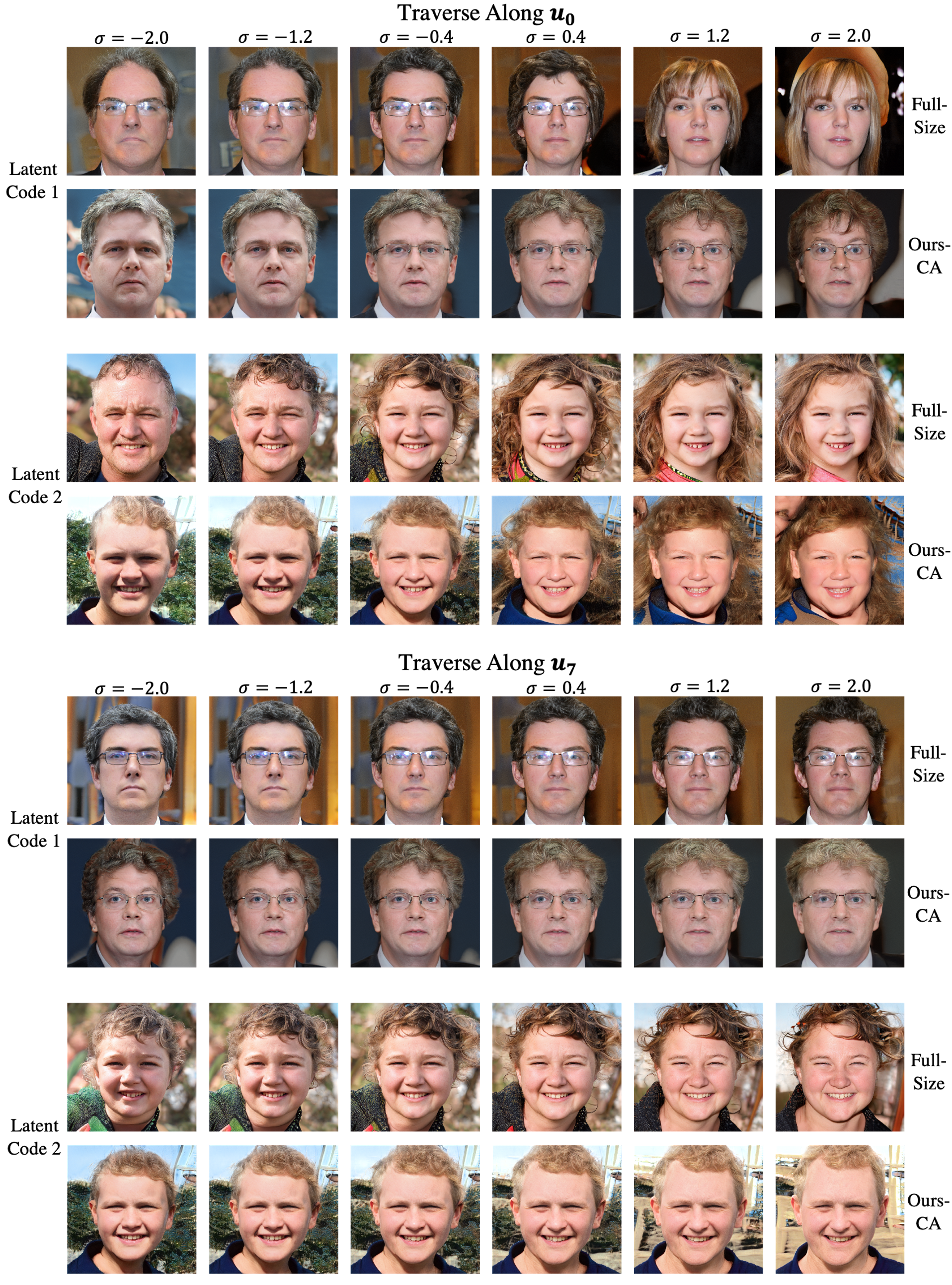}
    \vspace{-0.35cm}
    \caption{Additional results demonstrating the disentangled GANSpace editing directions with our compressed model.
    We traverse across $\mathbf{u}_0$ and $\mathbf{u_7}$ with two latent codes by the full-size model (74.3B FLOPs) and our compressed model (7.1B FLOPs).
    In $\mathbf{u}_0$ direction for gender editing, the full-size model again change the people's ages significantly while our model keeps them similar.
    Full-size model not only change size of the face and the hair with $\mathbf{u}_7$ direction, but also adjust people's identity undesirably.
    In contrast, our model preserves the identity well in the editing sequence.
    These results confirm again that our model enjoys a more disentangled latent manifold.}
    \label{fig:ganspace_extra}
\end{figure*}

We show additional examples of latent space image morphing in Fig.~\ref{fig:image_morphing} and GANSpace~\cite{harkonen2020ganspace} editing in Fig.~\ref{fig:ganspace_extra}.
We observe a smoother latent manifold and more disentangled editing directions in our compressed model.

\end{document}